\documentclass[10pt,twocolumn,letterpaper]{article}

\usepackage[accsupp]{axessibility}  

\usepackage{authblk}
\usepackage{cvpr}              

\usepackage{graphicx}
\usepackage{amsmath}
\usepackage{amssymb}
\usepackage{booktabs}
\usepackage{multirow}
\usepackage{algorithm2e}
\usepackage{adjustbox}
\usepackage{svg}
\usepackage{soul} 

\usepackage[pagebackref,breaklinks,colorlinks]{hyperref}

\usepackage[capitalize]{cleveref}
\crefname{section}{Sec.}{Secs.}
\Crefname{section}{Section}{Sections}
\Crefname{table}{Table}{Tables}
\crefname{table}{Tab.}{Tabs.}

\usepackage{color}

\def\cvprPaperID{9611} 
\def\confName{CVPR}
\def\confYear{2023}

\begin{document}

\title{NIPQ: Noise proxy-based Integrated Pseudo-Quantization}


\author[2]{Juncheol Shin\thanks{Equal contribution.}}
\newcommand\CoAuthorMark{\footnotemark[\arabic{footnote}]}
\author[1]{Junhyuk So\protect\CoAuthorMark}
\author[2]{Sein Park}
\author[3]{Seungyeop Kang}
\author[3]{Sungjoo Yoo}
\author[1,2]{Eunhyeok Park}
\affil[1]{Department of Computer Science and Engineering, POSTECH}
\affil[2]{Graduate School of Artificial Intelligence, POSTECH}
\affil[3]{Department of Computer Science and Engineering, Seoul National University \protect\\
{\tt\small \{jchshin, junhyukso, seinpark, eh.park\}@postech.ac.kr, \{ksy8653, yeonbin\}@snu.ac.kr }}
\renewcommand\Authands{, and }

\maketitle


\begin{abstract}
Straight-through estimator (STE), which enables the gradient flow over the non-differentiable function via approximation, has been favored in studies related to quantization-aware training (QAT). However, STE incurs unstable convergence during QAT, resulting in notable quality degradation in low precision. Recently, pseudo-quantization training has been proposed as an alternative approach to updating the learnable parameters using the pseudo-quantization noise instead of STE. In this study, we propose a novel noise proxy-based integrated pseudo-quantization (NIPQ) that enables unified support of pseudo-quantization for both activation and weight by integrating the idea of truncation on the pseudo-quantization framework. NIPQ updates all of the quantization parameters (e.g., bit-width and truncation boundary) as well as the network parameters via gradient descent without STE instability. According to our extensive experiments, NIPQ outperforms existing quantization algorithms in various vision and language applications by a large margin.
\end{abstract}


\section{Introduction}


Neural network quantization is a representative optimization technique that reduces the memory footprint by storing the activation and weight in a low-precision domain. In addition, when hardware acceleration is available (e.g., low-precision arithmetics~\cite{NVIDIA_8bit,song20197,NVIDIA_4bit,IntArithmetic} or bit-serial operations~\cite{Andrew,BitSerial,APNN}), it also brings a substantial performance boost. These advantages make network inference affordable in large-scale servers as well as embedded devices~\cite{server,phone}, which has helped popularize it in various applications. However, quantization has a critical disadvantage, quality degradation due to limited degrees of freedom. The way to train the networks accurately within limited precision is critical and receiving much attention these days.


To mitigate the accuracy degradation, quantization-aware training (QAT) has emerged that trains a neural network with quantization operators to adapt to low precision. While the quantization operator is not differentiable, the straight-through estimator (STE)~\cite{STE} allows the backpropagation of the quantized data based on linear approximation~\cite{DoReFa,DSQ}. This approximation works well in redundant networks with moderate precision ($>$4-bit). Thus, not only early studies~\cite{Courbariaux,XNOR,DoReFa} but also advanced ones~\cite{PACT,LSQ,DJPQ} have proposed diverse QAT schemes based on STE and shown that popular neural networks (i.e., ResNet-18~\cite{ResNet}) can be quantized into 4-bit without accuracy loss. 

\begin{figure}
    \centering
    \includegraphics[width=\columnwidth]{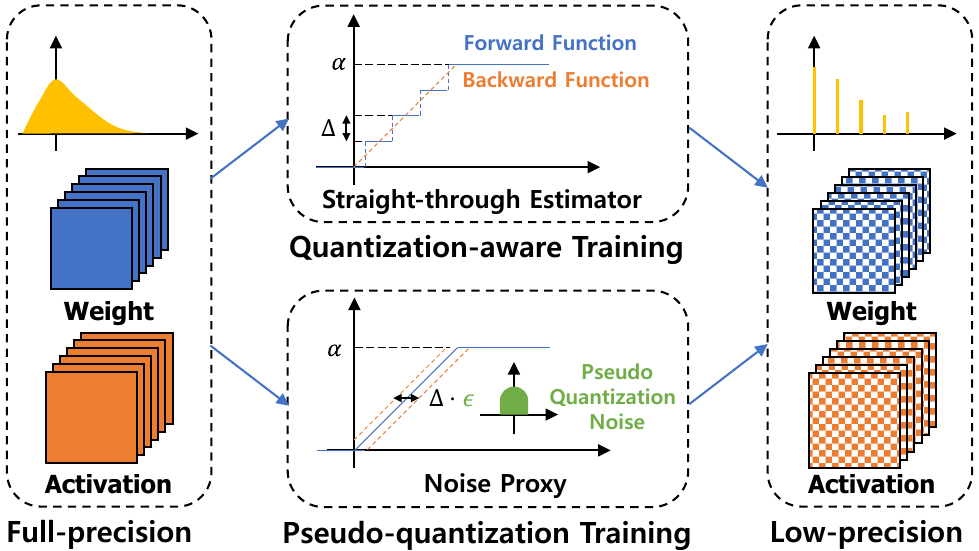}
    \caption{The proposed NIPQ as an alternative to QAT with STE.}
    \label{fig:overview}
\end{figure}


However, STE-based QAT bypasses the approximated gradient, not the true gradient, and many studies have pointed out that it can incur instability and bias during training~\cite{BiReal,DSQ,HLHLp,PROFIT,Mag, nagel2022overcoming}. 
For instance, PROFIT\cite{PROFIT} points out that the instability is a major source of accuracy drop for the optimized networks (e.g., MobileNet-v2/v3~\cite{MoV2,Mov3}). 
More recently, an alternative training scheme,  pseudo-quantization training (PQT) based on pseudo-quantization noise (PQN), has been proposed~\cite{NICE,DiffQ,NotAll} to address the instability of STE-based QAT. During PQT, the behavior of the quantization operator is simulated via PQN, and the learnable parameters are updated based on the proxy of quantization. While those studies are applied only to the weight, they can stabilize the training process significantly compared to QAT with STE and show the potential of PQT. 


Nevertheless, the existing PQT algorithms have room for improvement. Various STE-based studies~\cite{DoReFa,PACT,LSQ} have shown that truncation contributes significantly to reducing quantization errors, but even the advanced PQT studies~\cite{DiffQ,NotAll} use a naive min-max quantization. Integrating the truncation on the PQT framework can greatly reduce quantization error and enable us to exploit a PQT scheme for activation, which has an input-dependent distribution and requires a static quantization range. 
In addition, there is no theoretical support for whether PQT guarantees the optimal convergence of the quantization parameters. Intuitive interpretation exists, but proof of whether quantization parameters are optimized after PQT has yet to be provided. 



In this paper, we propose an novel PQT method (Figure \ref{fig:overview}) called Noise proxy-based Integrated Pseudo-Quantization (NIPQ) that quantizes all activation and weight based on PQN. We present a novel idea, called Noise proxy, that shares the same quantization hyper-parameters (e.g., truncation boundary and bit-width) with the existing STE-based algorithm LSQ(+)~\cite{LSQ,LSQ+}. However, noise proxy allows to update the quantization parameters, as well as the network parameters, via gradient descent with PQN instead of STE. Subsequently, an arbitrary network can be optimized in a mixed-precision representation without STE instability. Our key contributions are summarized as follows:
\begin{itemize}
  \item NIPQ is the first PQT that integrates truncation in addition to discretization. This extension not only further reduces the weight quantization error but also enables PQT for activation quantization. 
  \item NIPQ optimizes an arbitrary network into the mixed-precision with awareness of the given resource constraint without human intervention.
  \item We provide theoretical analysis showing that NIPQ updates the quantization hyperparameters toward minimizing the quantization error.   
  \item We provide extensive experimental results to validate the utility of NIPQ. It outperforms all existing mixed-precision quantization schemes by a large margin.  
\end{itemize}

\section{Related Work}

Due to its practical usefulness, various studies have been proposed for multi-bit linear quantization~\cite{DoReFa,PACT,QIL,LSQ}. Most of these studies are based on STE, and the proposed schemes have evolved by designing the quantization function to enable the optimization of the quantization parameters via gradient descent. However, in optimized networks such as MobileNet-v2, accuracy loss induced by STE instability has been reported for both activation~\cite{DSQ} and weight~\cite{PROFIT}. They addressed the instability via a newly designed pipeline and non-linear approximation but suffered from the increased complexity and cost of QAT. In this work, NIPQ updates the quantization parameters without STE approximation, enabling stable convergence without additional cost or complexity and, most importantly, outperforming the existing methods by a large margin.


Mixed-precision studies focus on allocating layer-wise or group-wise bit-width in consideration of the precision sensitivity of each layer to minimize accuracy drop within the given resource constraints. Various methods have been proposed, e.g., RL-based~\cite{HAQ,RELEQ,SAQ}, Hessian-based~\cite{HAWQ,HAWQv2,HAWQv3}, and differentiable~\cite{MPDNN,DJPQ} algorithms, but RL-based and Hessian-based methods are relatively complex, requiring a lot of parameter tuning, and differentiable algorithms still suffer from STE approximation error. In the present work, we reinterpret the differentiable bit-width tuning in terms of PQT framework, enabling the layer-wise bit-width tuning via gradient descent without STE instability. 


Robust quantization~\cite{KURE,Grad,Smooth} aims to guide the convergence of the network toward a smooth and flat loss surface based on additional regularization. The robustness of neural networks is highly beneficial for deploying noisy devices or low-precision ALUs. In the case of noise proxy, it inherently improves the robustness during PQT with PQN. As far as we know, we observe for the first time that the robustness of activation quantization can be enhanced (Section \ref{sec:robust}).


The benefit of the PQT has been demonstrated in diverse studies~\cite{DiffQ,NotAll} in various perspectives. However, previous studies utilize PQT with PQN in a limited domain only for weight and have yet to provide theoretical support for the convergence of PQT corresponding to the quantization. We extend the idea of PQT to both activation and weight by integrating the idea of truncation in addition to discretization and prove that the optimization of quantization parameters through noise proxy minimizes the actual quantization error. In addition, our work is the first to demonstrate that the PQT-based pipeline outperforms STE-based mixed-precision quantization algorithms by a large margin.

\section{Simple Example Problem}
Before introducing the details, we first propose a simple problem for straightforward explanation. The objective is to minimize the difference from the target data $t \in \mathbb{R}^N$ and the quantized value of the learnable parameters $x \in \mathbb{R}^N$. The example loss function is defined as follows:
\begin{equation}\label{eq:target}
    \mathcal{L}_{exp}^{Q}(x, \alpha, b) = \sum_{i=0}^{N-1}{\Big(t_i - Q(x_i | \alpha, b)\Big)^2},
\end{equation}
where subscript $i$ means the i-th element of the data, $t$ is arbitrary distribution, and $Q(\cdot)$ is a given quantization function, parameterized $\alpha$~(truncation boundary) and $b$~(bit-width). To minimize the loss, we need to update $x, \alpha$, and $b$ judiciously.
We utilize this example for the rest of the paper. 

\section{Motivation} \label{sec:motiv}
\begin{figure}
    \centering
    \includegraphics[width=1.0\columnwidth]{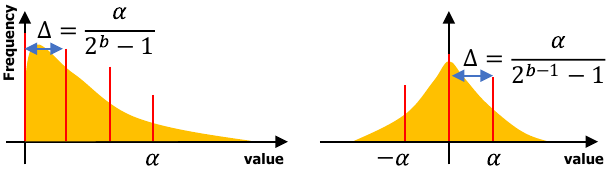}
    \caption{Visualization of the 2-bit quantization with hyper-parameters (e.g., $\alpha$ for truncation boundary and $b$ for bit-width) for non-negative distribution (left) and symmetric distribution (right) }
    \label{fig:quantization}
     \centering
     \includegraphics[width=1.0\columnwidth]{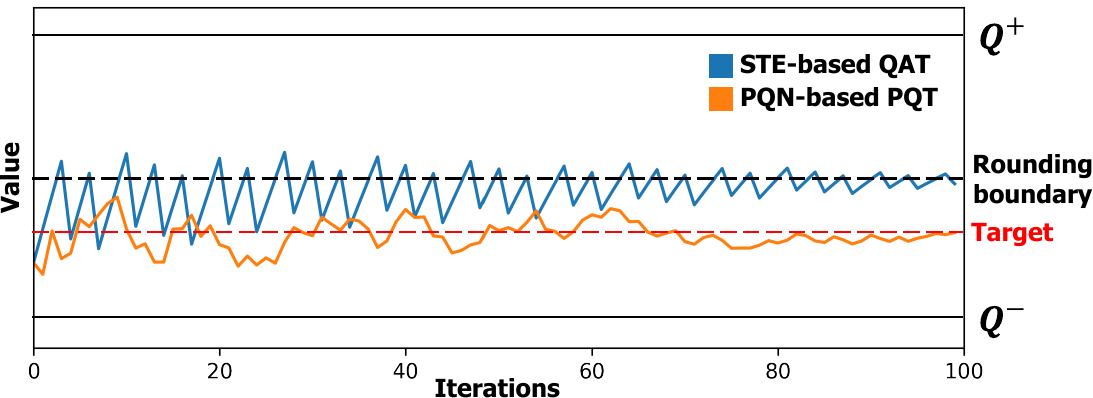}
     \caption{The trajectory of a scalar input during the minimization of $\mathcal{L}^Q$ through SGD with STE-based QAT and PQN-based PQT. $Q^+$ represents the positive quantization level and $Q^-$ the negative.}
     \label{fig:STE_vs_QAPT}
\end{figure}

In this work, we focus on linear quantization, especially for a more specific case that has an edge on hardware acceleration (e.g., affine layer-wise quantization for activation and symmetry channel-wise quantization for weight)~\cite{IntArithmetic}. To get optimal quality within restricted resources, we need to tune the quantization parameters, shown in Figure \ref{fig:quantization}. In this section, we analyze the limitations of existing STE-based and PQT-based algorithms and provide insight for next-step innovation. 

\subsection{Limitation of STE-based Quantization}
\label{sec:STElimit}
STE~\cite{STE} allows the gradient flow over the quantized data by approximating the discretization as linear identity mapping. STE-based QAT updates the quantization parameters via gradient descent, which enables the joint update of network and quantization parameters toward minimizing the target loss value. Many prior works show successful results relying on this approximation~\cite{DoReFa,DJPQ,LSQ,QIL,PACT,BiReal,PROFIT}.



However, STE-based QAT has a critical limitation. As shown in Figure \ref{fig:STE_vs_QAPT}, the scalar value never converges to the target value; instead, it keeps oscillating near the rounding boundary in the middle of the two nearest quantization levels. This is because when the scalar value is updated to cross the rounding boundary, it is mapped to a different quantization level. Then the sign of the gradient is reversed, updating the scalar value toward the opposite direction. The same process is repeated during training. Even at the later iterations, when the learning rate gradually decreases, the scalar value still oscillates near the rounding boundary.

The frequent flipping of the quantization value induces instability in network convergence. Moreover, the oscillation near the rounding boundary becomes the major source of large quantization errors. Even at the end of the training, $Q(x|\alpha, b)$ can be mapped to either $Q^+$ and $Q^-$, while $Q^-$ has a minimal difference from the target value. Many studies have pointed out that this phenomenon degrades the quality of the quantized network, especially for the optimized networks~\cite{PROFIT,Mag,nagel2022overcoming}.

\begin{figure}
    \centering
    \includegraphics[width=1.0\columnwidth]{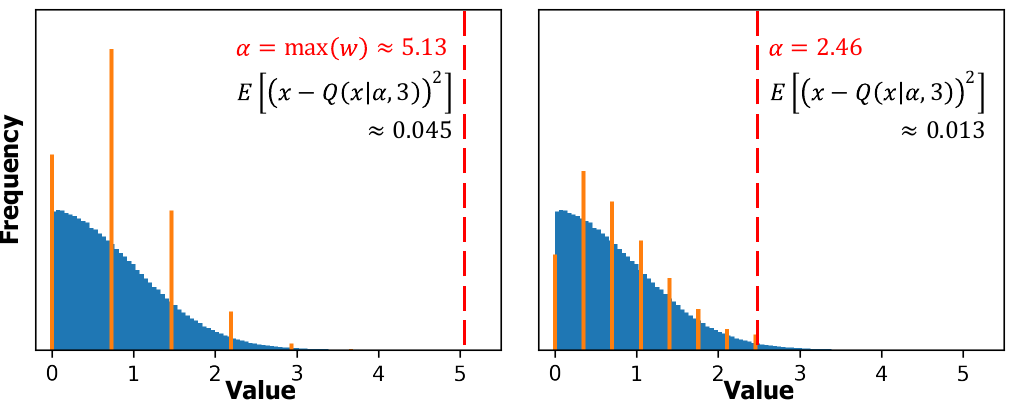}
    \caption{The comparison of 3-bit min-max quantization (left) and 3-bit quantization with truncation (right). }
    \label{fig:quanterror}
\end{figure}

\subsection{Pros and Cons of Previous PQN-based PQT}
To tackle the limitation of STE-based QAT, an alternative quantization pipeline, PQN-based PQT, has been proposed~\cite{DiffQ,NotAll}. During PQT, no quantization operator is used for forward and backward operation, but the behavior of the quantization operator is simulated via PQN. The learnable parameters are updated in a differentiable way without STE instability based on the proxy of quantization, becoming robust in the low-precision domain.
As shown in Figure \ref{fig:STE_vs_QAPT}, the scalar value with PQT is updated stably toward the target value with small oscillation.
Especially as the learning rate gets decreased, oscillation amplitude becomes reduced, eventually converging to the target value.
After the PQT, the scalar value has been firmly mapped to $Q^-$, leading to minimal quantization error.
Compared to STE-based QAT, PQN-based PQT is expected to have lower quantization error after the training.

However, existing PQT-related studies still have critical limitations. The advanced methods~\cite{DiffQ,NotAll} utilize min-max quantization, which uses the minimum/maximum value of data for the quantization range. Many STE-based studies point out that truncation is essential to minimize the quantization error in the low-precision domain~\cite{DoReFa,PACT,LSQ,QIL}.
As shown in Figure \ref{fig:quanterror}, applying quantization based on the minimum/maximum value of the data wastes invaluable quantization levels, resulting in higher quantization error.
As a long tail or outlier pushes the boundary by a large margin, truncation reduces overall quantization error significantly, especially for near-zero values.
In addition, quantization with truncation is crucial for extending PQT for activation quantization; because the activation is input-dependent, the static boundary needs to be trained over a training dataset for stable and fast inference.
Unfortunately, the existing studies do not provide the theoretical integration of truncation on top of a PQN-based PQT framework.

\section{NIPQ: Noise Proxy-based Integrated Pseudo-Quantization}


In this work, we propose a novel PQT pipeline to compensate for the quality degradation after quantization without STE instability. Moreover, the proposed scheme includes the support of updating truncation, resulting in accurate quantization not only for weight but also for activation.  Besides, we adopt the gradient-based bit-width optimization via continuous relaxation~\cite{DJPQ, DiffQ}. This extension allows us to update all of the hyper-parameters for linear quantization via gradient descent, which greatly simplifies the layer-wise optimization process. 

In this section, we define the proposed idea and present the analytical support for the optimal convergence of quantization parameters based on PQT. Our method is composed of a pair of algorithms: STE-based quantization $Q(x|\alpha,b)$~\cite{LSQ,LSQ+} and the corresponding noise proxy $\tilde{Q}(x|\alpha,b)$. The algorithms are expressed as follows:  
\begin{align}
Q(x|\alpha,b) = \left\{\begin{matrix}
0, & x \le 0 \\ 
\lfloor x / \Delta\rceil \cdot \Delta, & 0 < x < \alpha\\
\alpha, & x \ge \alpha, 
\end{matrix}\right. \label{eq:q}
\\
\tilde{Q}(x|\alpha,b) = \left\{\begin{matrix}
0, & x \le 0 \\ 
x + \epsilon\cdot\Delta, & 0 < x < \alpha\\
\alpha, & x \ge \alpha,
\end{matrix}\right. \label{eq:qnoise}
\end{align} 
where  $\Delta = \frac{\alpha}{2^b-1}$ represents the quantization step size or the difference between quantization levels, $\epsilon$ represents the PQN, and $\lfloor \cdot \rceil$ means the rounding operation\footnote{Please note that we present the quantization algorithm for non-negative distribution data, e.g., activation after ReLU, for simplicity. However, this approach holds the same for weight, except that quantization index is used in $[-2^{bit-1}, 2^{bit-1}-1]$.}. 

During PQT, $\tilde{Q}(x|\alpha, b)$ is applied on top of the quantization targets, activation and weight, and the learnable parameters are updated via gradient descent. Because $\alpha$ and $ b$ are included in the gradient path, the training with noise proxy updates all of the quantization parameters as well as the network's learnable parameters jointly.
If we set the target loss function, $\mathcal{L}_{target}$ as the mixture of the task loss $\mathcal{L}_{task}$, and the cost loss $\mathcal{L}_{cost}$, for memory footprint or computation as follows:
\begin{equation}
    \mathcal{L}_{target} = \mathcal{L}_{task} +  \lambda \cdot \mathcal{L}_{cost}, 
\end{equation}
where $\lambda$ is a hyper-parameter of balancing two loss functions. The layer-wise bit-width $b$ is assigned within the available resource budget, and the corresponding $\alpha$ is tuned appropriately to achieve the best quality possible.


After PQT, $Q(x|\alpha, b)$ is used for the true quantization during inference. 
To achieve the high-quality output with the true quantization operator, the tuned $\alpha$ and $b$ through PQT with noise proxy should also have the lowest loss with $Q(x|\alpha, b)$. In the following subsections, we provide theoretical support that the optimization of $\alpha$ and $b$ through PQT guarantees the optimal convergence $Q(x|\alpha, b)$ when we sample PQN judiciously.



\subsection{Input Activation Convergence}\label{sec:act_converge}
First of all, we evaluate the convergence property of the learnable input tensor $x$. In STE-based algorithm, the gradient of input data is bypassed through STE algorithm. Thereby the derivative is expressed as follows:
\begin{equation}
    \frac{\partial Q(x|\alpha,b)}{\partial x} = \left\{\begin{matrix}
0, & x \le 0\\ 
1, & 0 < x < \alpha\\
0, & x \ge \alpha.
\end{matrix}\right.
\end{equation}
In the case of the example in Eq.~\ref{eq:target}, the gradient of $x_i$ within the quantization interval ($x_i \in [0, \alpha]$) is calculated as follows:
\begin{equation}
    \frac{\partial \mathcal{L}^{Q}_{exp}(x, \alpha, b)}{\partial x_i} = 2(Q(x_i|\alpha,b) - t_i).
\end{equation}

Therefore, $x_i$ converges to $t_i$ theoretically as the training progresses. However, as shown in Figure \ref{fig:STE_vs_QAPT}, $x_i$ in practice fluctuates near the rounding boundary, except $Q(x_i|\alpha,b) = t_i$ exist, due to the instability of STE-based QAT.

On the other hand, the gradient of input data regarding noise proxy is expressed as follows:
\begin{equation}
    \frac{\partial \tilde{Q}(x|\alpha,b)}{\partial x} = \left\{\begin{matrix}
0, & x \le 0\\ 
1, & 0 < x < \alpha\\
0, & x \ge \alpha,
\end{matrix}\right.
\end{equation}
which is exactly the same as the $Q(\cdot)$ condition. When we update the example loss function via the proposed algorithm, the gradient of $x_i$ within the quantization interval is evaluated as follows:
\begin{equation}
    \frac{\partial \mathcal{L}^{\tilde{Q}}_{exp}(x, \alpha, b)}{\partial x_i} = 2(\tilde{Q}(x_i|\alpha,b) - t_i) = 2(x_i+\epsilon\cdot\Delta-t_i).
\end{equation}

The gradient of $x_i$ points toward the target, $t_i$, but with drift and noise from PQN. However, the drift can be amortized when the average of PQN is zero-centered ($E[\epsilon] = 0$). Note that the distribution of actual quantization errors is empirically known to have zero-mean~\cite{PQN}. In addition, because the update step size is proportional to the learning rate, the oscillation amplitude becomes smaller when the learning rate becomes smaller as learning progresses; eventually, $x_i$ converges to the target value, as shown in Figure \ref{fig:STE_vs_QAPT}. Unlike the STE-based QAT, noise proxy-based PQT precisely evaluates the effect of oscillation induced by PQN and discards the bias of PQN gradually during training.


Besides, the convergence property of oscillation near the target value is greatly beneficial to enhance the robustness of network, as pointed out in several previous studies~\cite{Smooth}. When some noise, $\delta$, exists, the objective function can be  approximated via Taylor expansion as follows: 
\begin{align}
    &E[\mathcal{L}(x+\delta)] \\ &\approx E[\mathcal{L}(x)+\delta \cdot \nabla_x \mathcal{L}(x) + \frac{1}{2}\delta^T\cdot\nabla_x^2\mathcal{L}(x)\cdot\delta]\\
    & \approx \mathcal{L}(x) + \frac{E[\delta^2]}{2}Tr\Big\{\nabla_x^2\mathcal{L}(x)\Big\},
\end{align}
where the term related to the first derivative is removed since $E[\delta] = 0$, and the off-diagonal elements of the second derivative term become 0 because it relates to the expectation of multiplication of two i.i.d. samples. When the loss is converged to the minima point, the sum of eigenvalues, $\nabla_x^2L(x)$, should have a non-negative value. To minimize the average loss after training, the loss surface is guided to be converged to the minima having a lower Hessian trace value~\cite{Smooth}, resulting in enhanced network robustness.




\subsection{Alpha and Bit-width Convergence}
In addition to the input data, the noise proxy should gurantee the convergence of the quantization parameters in the optimal point that minimizes the loss with $Q(x|\alpha, b)$. First, let's consider the gradient of $\alpha$ regarding quantization and noise proxy function. 
\begin{align}
\frac{\partial Q(x|\alpha,b)}{\partial \alpha} = \sum_{x_i\ge\alpha}{1} + \sum_{0<x_i<\alpha}{\frac{1}{N_{lv}}\lfloor\frac{N_{lv}}{\alpha}x\rceil-\frac{x}{\alpha}}, \label{eq:dela1}
\\
\frac{\partial \tilde{Q}(x|\alpha,b)}{\partial \alpha} = \sum_{x_i\ge\alpha}{1} + \sum_{0<x_i<\alpha}{\epsilon\cdot\frac{1}{N_{lv}}}, \label{eq:dela2}
\end{align}
where $N_{lv}=2^{b}-1$ denotes the number of levels. 

Two gradients have a difference only for the case when $x \in (0, \alpha)$. However, note that Equations \ref{eq:dela1} and \ref{eq:dela2} become identical when PQN $\epsilon$ is sampled from the quantization noise distribution $\lfloor x/\Delta\rceil-x/\Delta$. The same conclusion is observable for the number of quantization levels, which is expressed as:
\begin{align}
\frac{\partial Q(x|\alpha,b)}{\partial N_{lv}} = \sum_{0<x_i<\alpha}{\frac{x}{N_{lv}}-\frac{\alpha}{{N_{lv}}^2}\lfloor\frac{N_{lv}}{\alpha}x}\rceil,
\\
\frac{\partial \tilde{Q}(x|\alpha,b)}{\partial N_{lv}} = \sum_{0<x_i<\alpha}{-\epsilon\frac{\alpha}{N_{lv}^2}}.
\end{align}
Therefore, when we use the sampled noise following the quantization error distribution, the gradient of quantization parameters becomes identical to the true quantization operator while still enjoying the benefit of stochasticity of PQN. 
\textcolor{blue}{}

In short, when we apply NIPQ whose noise is sampled from the quantization noise distribution, the gradient direction of $\alpha$ and $b$ is identical in both $Q(x|\alpha,b)$
and $\tilde{Q}(x|\alpha,b)$. Therefore, when $\alpha$ converges to the optimal point via PQT where $\frac{\partial\mathcal{L^{\tilde{Q}}}}{\partial\alpha} = \Sigma_i\frac{\partial L}{\partial \tilde{Q}_i(x_i|\alpha,b)}\cdot
\frac{\partial \tilde{Q}_i(\cdot)}{\partial \alpha} = 0$, then we can easily validate that $\frac{\partial\mathcal{L^{Q}}}{\partial\alpha}=\Sigma_i\frac{\partial L}{\partial Q_i(x_i|\alpha,b)}\cdot
\frac{\partial Q_i(\cdot)}{\partial \alpha} = 0$. Likewise, when $\frac{\partial\mathcal{L^{\tilde{Q}}}}{\partial b}=0$, then $\frac{\partial\mathcal{L^{Q}}}{\partial b}=0$ and that is the point where the loss value with $Q(x|\alpha,b)$ is minimized. Therefore, optimizing the quantization parameters $\alpha$ and $b$ based on NIPQ also minimizes the target loss with true quantization.

\section{Additional Details}
As explained above, NIPQ enables the compensation of quality degradation of the quantized network without STE approximation. However, in practice, we need to consider several details to maximize obtainable accuracy.

\subsection{Stochastic Rounding for Bit-width Parameter}
First, during PQT phase, we use the bit-width with stochastic rounding as follows: 
\begin{align}
    b \leftarrow 2 + 14\cdot{Sigmoid(\beta)}, \label{eq:b1} \\
    b \leftarrow \lfloor b + U(-0.5, 0.5) \rceil \label{eq:b2},
\end{align}
where $U(x, y)$ represents the uniform noise in $[x, y]$, and $\beta$ is a trainable value in the continuous domain corresponding to the bit-with $b$. The bit-with is narrowed into a range of 2 to 16 bits, then mapped to the integer value via stochastic rounding. The idea of continuous relaxation of bit-width and updating it through gradient descent has been proposed in several previous studies~\cite{DiffQ,DJPQ}. They have used the continuous value as it is, but we find that these settings yield suboptimal convergence because of the domain difference in that only discrete bit-width can be a candidate of precision during inference. To resolve the domain gap, we propose the bit-width update with stochastic rounding. It is an unbiased estimator but makes a decision in discrete space, whereby the domain difference could be mitigated. Our experiment shows that this update improves the final accuracy by a large margin, especially in the sub-4-bit domain. The experiments are included in the Supplementary Materials. 

\subsection{Stabilization in the Late Training Stage}
In addition, we need to update the batch normalization layers after finishing NIPQ training (BN update). PQN in NIPQ induces instability in the running statistics of the normalization layers, so we update the batch normalization statistics with the true quantization operator after finishing training. We also observe additional performance improvement by switching from NIPQ to STE-based QAT at the later stage of training and updating the last few epochs with a very small learning rate (QAT finetune), while the amount of improvement is far smaller than BN update. We speculate that QAT finetune gives an opportunity to finetune the learnable parameters as well as normalization statistics of the network to mitigate the minor instability form noise. The details are included in the Supplementary Materials. 

\subsection{PQN Sampling}
Finally, we provide the convergence condition in the last section in that PQN should be sampled following the quantization error distribution. However, we observe that NIPQ with a uniform noise $U(-\frac{\Delta}{2},\frac{\Delta}{2})$ also shows the comparable result to the noise with the desired property\footnote{Unlike the previous studies~\cite{DiffQ,NICE}, PQN from normal distribution shows inferior performance in our algorithm.}. In this configuration, the relative frequency of the quantization error is ignored. This makes the quantization noise overestimated, and the low bit tends not to be selected, but a similar result is achievable by making $\lambda$ of the cost function larger. In particular, the difference in final quality is negligible if the QAT finetune is applied. In practice, sampling of quantization error distribution slows down the training significantly. Uniform noise with QAT finetune can be considered as a practical approach of reducing PQN sampling overhead while exploiting the benefit of PQT.

\section{Experimental Setup}
In order to validate the performance of NIPQ, we conduct comprehensive studies to measure the output quality and observe diverse properties in various applications. The results in this paper are obtained based on PQN sampled from the quantization noise, unless otherwise specified explicitly. Likewise, QAT finetune is used at the last few epochs, otherwise specified explicitly. We implement our algorithm on PyTorch~\cite{PyTorch} library, and the code is available at our repository~\footnote{https://github.com/ECoLab-POSTECH/NIPQ}. The details of target loss and training pipeline are available in the Supplementary Materials.

\begin{figure}[t]
    \begin{subfigure}{\columnwidth}
         \includegraphics[width=1\textwidth]{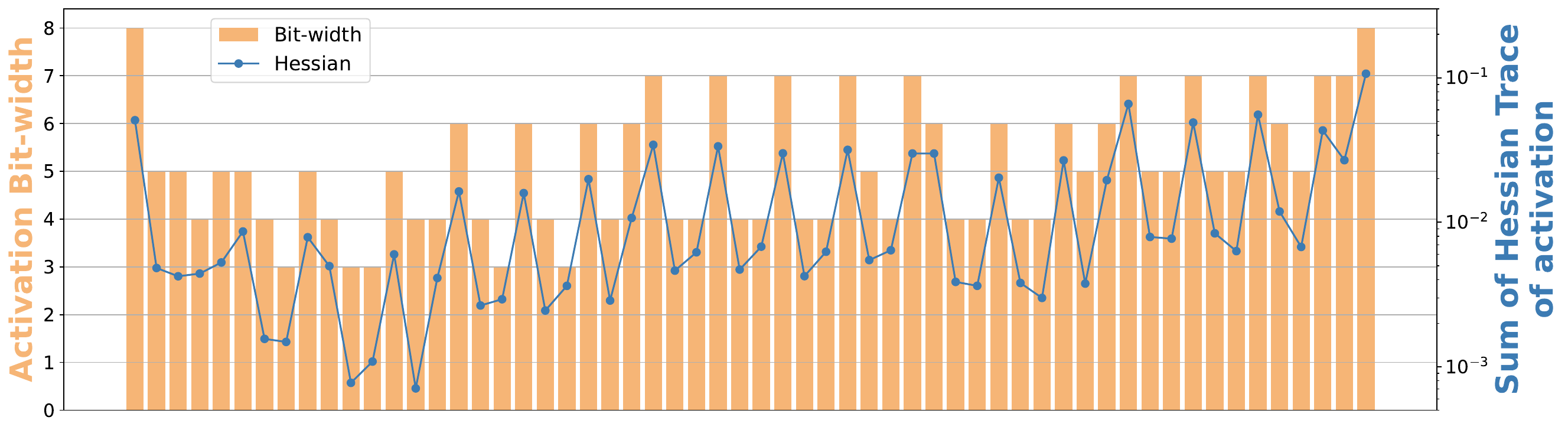}
     \end{subfigure}
     \begin{subfigure}{\columnwidth}
         \includegraphics[width=1\textwidth]{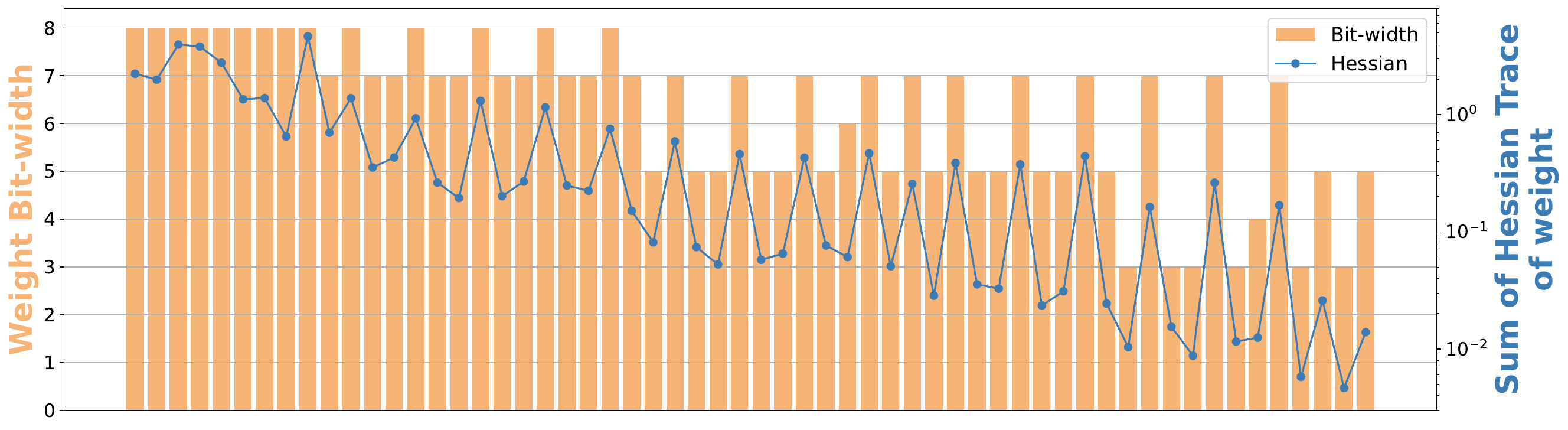}
     \end{subfigure}
     \rule[1ex]{\columnwidth}{0.1pt}
     \begin{subfigure}{\columnwidth}
         \includegraphics[width=1.0\textwidth]{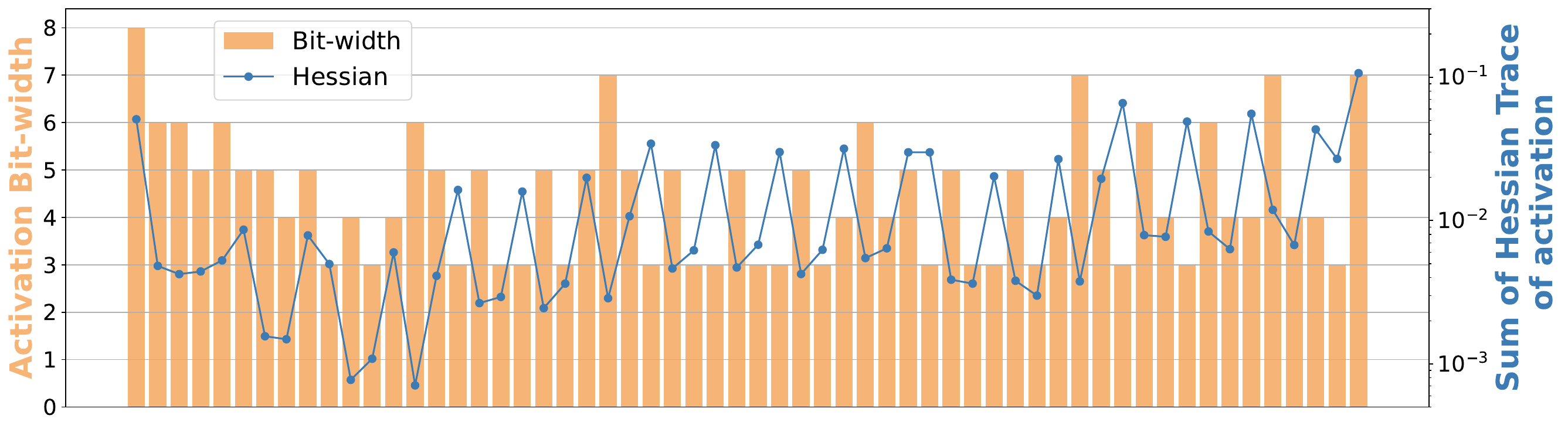}
     \end{subfigure}
     \begin{subfigure}{\columnwidth}
         \includegraphics[width=1.0\textwidth]{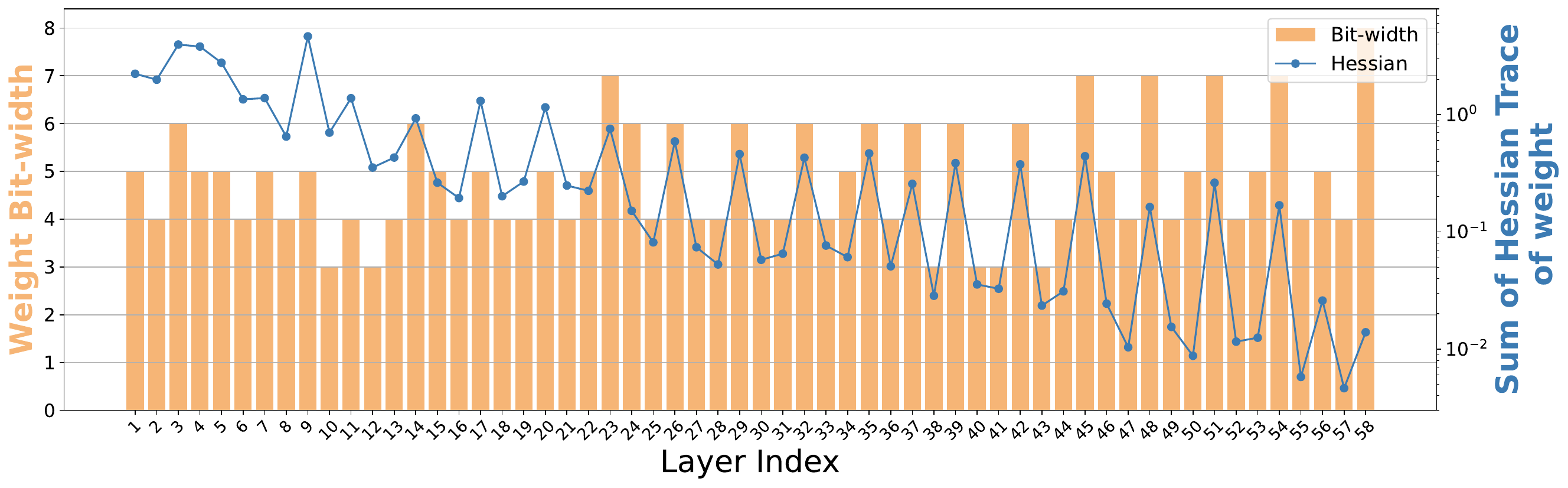}
     \end{subfigure}
     \caption{Layer-wise assigned bit-width of activation (top) and weight (bottom) and corresponding sensitivity measure (the sum of Hessian trace) of MobileNet-v2 on the CIFAR-100 dataset~\cite{CIFAR}. The average precision of activation weight is restricted to 4-bit (top) and the computation is restricted to 1.5 GBOPs (bottom).  
     }
     \label{fig:bit-hessian}
\end{figure}

\subsection{Sensitivity-aware Mixed-precision}



Mixed-precision quantization aims at improving accuracy while minimizing the overall storage footprint and computational cost. To maintain accuracy within the resource budget, more bit-width should be assigned to the sensitive layer to minimize overall quantization errors. Recently, the Hessian of the loss function has often been used as a metric of sensitivity against quantization~\cite{HAWQ,HAWQv2,HAWQv3}. The smaller the sum of the Hessian trace, the lower the sensitivity, and vice versa. Figure \ref{fig:bit-hessian} (top) visualizes the assigned bit-width of activation and weight and the corresponding sum of the Hessian trace estimated via the Hutchinson algorithm~\cite{HAWQv2} when applying NIPQ to MobileNet-v2 on the CIFAR-100 dataset. Each plot is obtained by optimizing the weights and activation independently. As shown in the figure, the more sensitive the layer is, the more bit-width is assigned. The redundant layers tend to have a fewer bit-width to match the target objective. Note that NIPQ does not have any additional stages that measure the sensitivity of the layer. Instead, we just train the network with an additional penalty term to restrict the average precision to 4-bit. The noise proxy algorithm allocates precision by itself, considering the sensitivity of the target layer, thereby quantizing the network with the highest accuracy as efficiently as possible within the constraints. 


On the other hand, the automated bit-width allocation enables tuning of network considering complex metric that is practically improbable with the conventional Hessian-based methods~\cite{HAWQ,HAWQv2,HAWQv3}.
Figure \ref{fig:bit-hessian} (bottom) shows the assigned bit-width of activation and weight when restricting the computation cost (bit-operations, BOPs~\cite{uniq,PROFIT}) as a 1.5G BOPs, which is equal to the computation cost of the quantized 4-bit model using PACT~\cite{PACT} or LSQ~\cite{LSQ}. Unlike Figure \ref{fig:bit-hessian} (top), the bit-width of activation is slightly misaligned with the sum of the hessian trace. In Figure \ref{fig:bit-hessian} (bottom), we aim to optimize the average bit-width of activation and weight independently. Those two configurations are optimized via disjoint target losses thereby, each precision is assigned proportionally to the sensitivity of activation and weight separately. However, in the case of BOPs, the computation cost is proportional to the product of bit-width of activation and weight. Thereby, when we restrict the overall computation cost, the activation bit-width is allocated with awareness of the activation sensitivity as well as the corresponding weight sensitivity, and vice versa. This experimental result indicates that assigning the bit-width based on layer-wise sensitivity naively might not be an optimal policy for computation-aware quantization. 

\begin{figure}[t]
     \centering
     \begin{subfigure}[b]{0.22\textwidth}
         \centering
         \includegraphics[width=\textwidth]{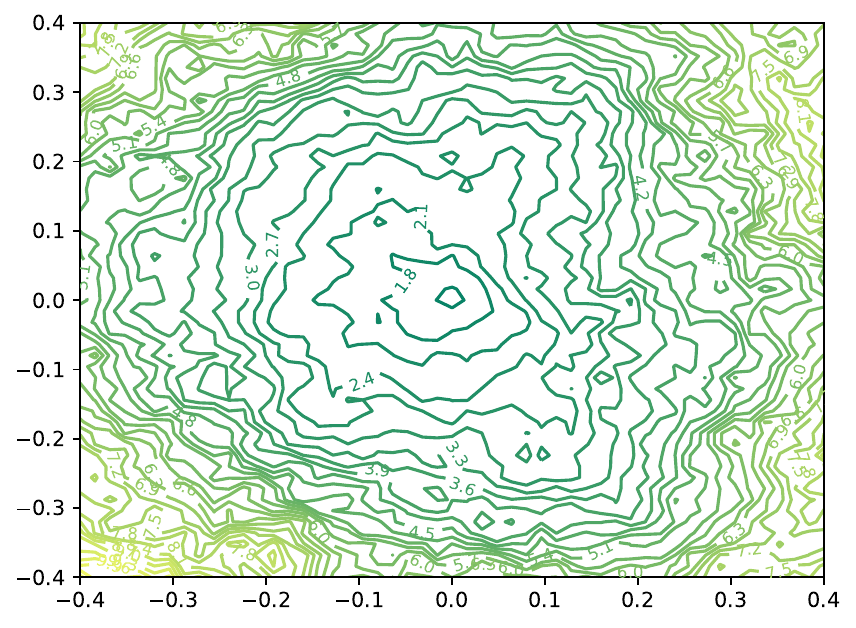}
     \end{subfigure}
    \begin{subfigure}[b]{0.22\textwidth}
         \centering
         \includegraphics[width=\textwidth]{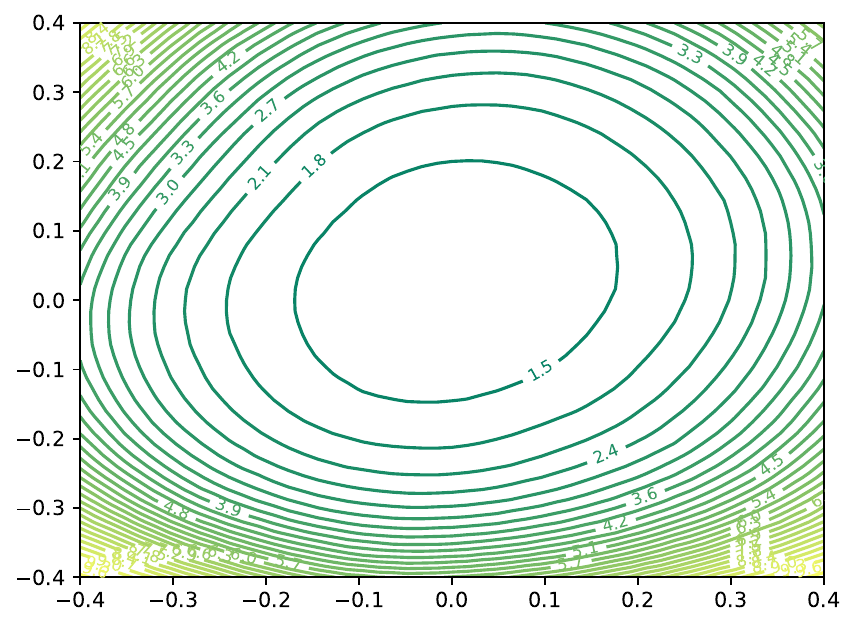}
     \end{subfigure}
     \begin{subfigure}[b]{0.22\textwidth}
         \centering
         \includegraphics[width=\textwidth]{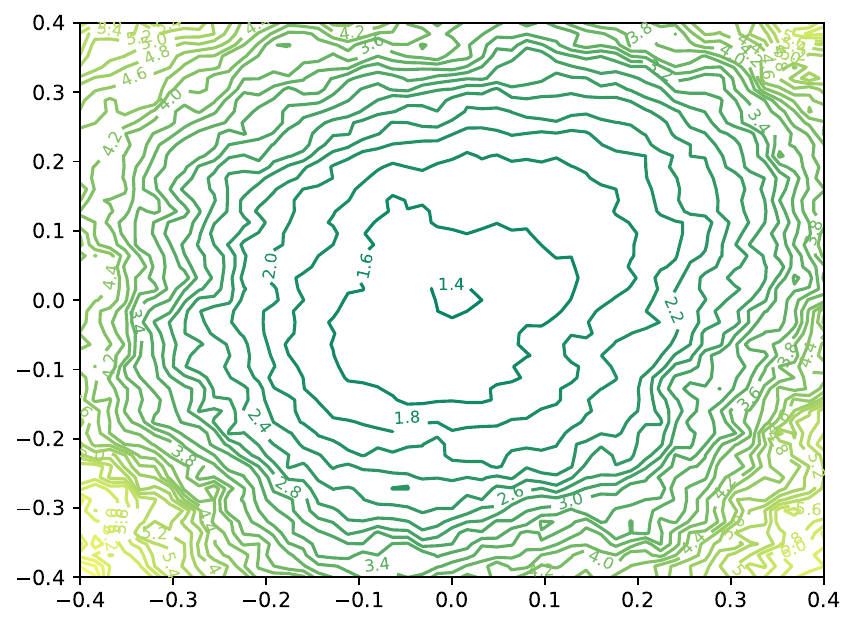}
     \end{subfigure}
     \begin{subfigure}[b]{0.22\textwidth}
         \centering
         \includegraphics[width=\textwidth]{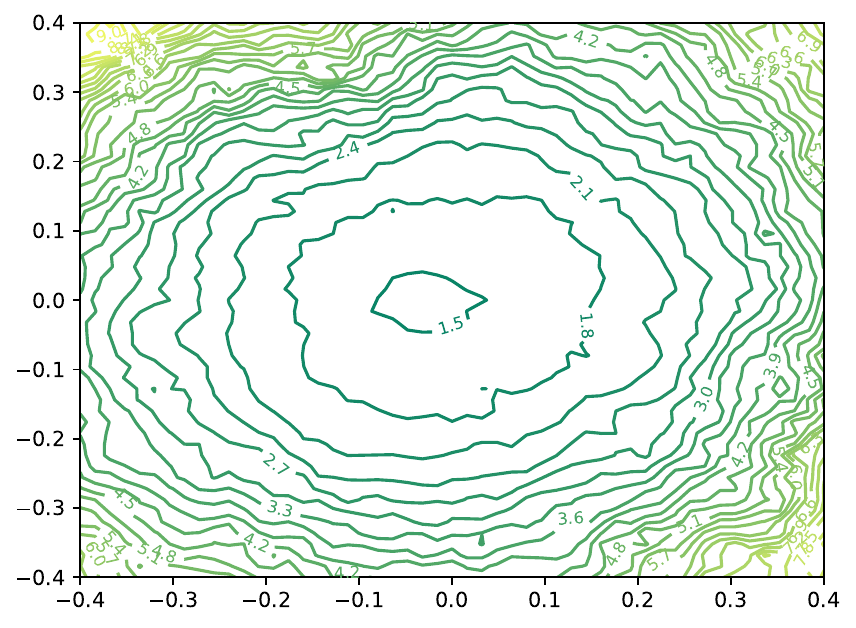}
     \end{subfigure}
     \caption{Loss landscapes~\cite{Viz} of 3-bit quantized MobileNet-V2 fine-tuned by STE-based LSQ~\cite{LSQ} (top left), 3-bit NIPQ, BN update (top right),  mixed-precision NIPQ (avg. 3-bit), BN update (bottom left), mixed-precision NIPQ (avg. 3-bit), QAT finetune (bottom right) on the CIFAR-100 dataset.}
     \label{fig:losssurf}
\end{figure}

\subsection{Robust Quantization for Practical Deployment}\label{sec:robust}

Another strength of NIPQ is the enhanced robustness of the network against unexpected noise. This brings diverse advantages to deploying the network in practice. For instance, the analog sum-based device~\cite{ISAAC} has significant energy efficiency but has inevitable noise induced by process variation or temperature drift, which causes instability of output. Even in the case of digital NPU, the low-precision implementation is fragmented because there are hundreds of hardware manufacturers~\cite{KURE}. When we deploy a neural network to the target device, the implementation difference can introduce unexpected noise on the data, resulting in accuracy degradation. The robustness of the network allows accuracy to be maintained in this environment, so securing this property is a significant advantage.


As explained in Section \ref{sec:act_converge}, NIPQ regularizes the loss surface. Figure \ref{fig:losssurf} visualizes the sharpness of the loss surface by measuring the change of loss values after adding noise on top of the trained weight along the two random vectors~\cite{Viz}. As shown in the figure, NIPQ converges to a flat and smooth loss surface (top left vs. right). When layer-wise mixed precision is enabled, the loss surface becomes complex (top right vs. bottom left). We speculate that mixed-precision optimization exploits the sensitivity to trade robustness for better output quality. NIPQ enables such a trade-off in an automatic manner by forcing the network to adapt to the pseudo-noise that generalizes the quantization noise. 

Besides, NIPQ enhances the robustness of the quantization parameters as well as the network parameters. Check the Supplementary Materials for the detailed results.

\subsection{Importance of Truncation for Quantization} \label{sec:vsdiffq}

\begin{table}[t]    
    \centering
    \begin{adjustbox}{{width=\columnwidth}}
    \begin{tabular}{ccccccccc}
    \toprule
            & MNLI  & CoLA  & MRPC  & QNLI  & QQP   & RTE   & SST-2 & STS-B     \\
    \cline{2-9}
     FP & 84.46 & 56.53 & 90.62 & 90.66 & 87.76 & 71.19 & 92.32 & 88.64 \\
     \hline     
     DiffQ-MP-6\cite{DiffQ}    & 81.66     & 0.04     & 85.85 & 50.54 & 86.65     & 47.29     & 89.91    & 74.86      \\
     NIPQ-MP-6     & \textbf{83.87}     & \textbf{59.60}     & \textbf{89.68}     & \textbf{90.92}     & \textbf{87.48}     & \textbf{68.95}         & \textbf{91.97}     & \textbf{88.11}       \\
     \hline
     DiffQ-MP-3\cite{DiffQ}    &   31.82   & 0.02     & 81.22     & 49.46 & 0    & 47.29 & 50.92 & 11.27          \\
     NIPQ-MP-3  & \textbf{81.88}     & \textbf{53.03} & \textbf{88.62} & \textbf{90.23} & \textbf{86.68} & \textbf{69.31} & \textbf{91.51} & \textbf{88.42}     \\
    \bottomrule
    \end{tabular}
    \end{adjustbox}
    \caption{Results of weight quantization for BERT-base on GLUE}
    \label{tab:bert}    
\end{table}

The last line of the related study is DiffQ~\cite{DiffQ}, which is based on PQN-based PQT for weight quantization. The key difference from NIPQ is that DiffQ uses linear quantization based on the min-max value of weight instead of truncation. However, to minimize quantization errors with the limited bit-width, the presence of truncation is extremely helpful. In general, the data distribution of a natural network follows a bell-shaped distribution, where the majority of data is concentrated near zero. Because min-max quantization increases the quantization interval near zero, the quantization error greatly increases. Table \ref{tab:bert} compares the accuracy after the quantization between DiffQ and NIPQ. Note that only the weight is quantized, and bit-width is allocated per layer. As shown in the table, NIPQ outperforms DiffQ by a large margin in the same bit-width. Integration of truncation on quantization minimizes the overall quantization error significantly, especially when the precision is limited.

\subsection{Quantization of Large-scale Vision Tasks}\label{sec:vision}
To demonstrate the outstanding performance of the proposed method, we apply NIPQ for large-scale vision applications, including ImageNet~\cite{ImageNet} classification, multi-scale super-resolution, and VOC~\cite{VOC} object detection. 

First, we apply quantization to diverse networks on the ImageNet~\cite{ImageNet} classification task, and performs comparisons with various existing studies. Existing studies report mixed results with/without using well-known teacher-student-based knowledge distortion, so we report the accuracy for both cases, with EfficientNet-B0 as a teacher when necessary. Only the input image has 8-bit precision, and every layer in the network, including the first and last layers, is quantized via noise proxy. 


\begin{table}[t]    
    \centering
    \begin{adjustbox}{width=0.45\textwidth}
        \begin{tabular}{ccccc}
    \toprule
         Model &  Method &Bit(W/A) & BOPs(G)& Top-1\\
    \toprule
    \multirow{13}{*}{ResNet-18\cite{ResNet}}
     & FP & 32/32 & 1857.6 & 70.54 \\
     & FP+KD & 32/32 & 1857.6 & 72.17\\
    \cline{2-5}
     & PACT\cite{PACT} & 4/4 & 34.7 & 69.2\\
     & LSQ\cite{LSQ}& 4/4 & 34.7 & 69.39\\
     & DJPQ\cite{DJPQ} & MP-BOPs & 35.0 & 69.3\\
     & HAQ\cite{HAQ} & MP-BOPs & 34.4 & 69.2\\
     & HAWQ\cite{HAWQ} & MP-BOPs & 34.0 & 68.5\\
     & HAWQ-V3\cite{HAWQv3} & MP-BOPs & 34.0 & 68.5\\
     & HAWQ-V3\cite{HAWQv3} & MP-BOPs & 72.0 & 70.2\\
    \cline{2-5}
     & NIPQ-MP-4 & 3.89/3.98 & 45.5 & \textbf{69.84}\\
     & NIPQ-BOPs & 4.53/3.67 & 34.1 & \textbf{69.47}\\
     & NIPQ-MP-4+KD & 4.09/4.17 & 50.73 & \textbf{71.83}\\
     & NIPQ-BOPs+KD & 4.47/3.65 & 34.2 & \textbf{71.24} \\
    \toprule
    \multirow{11}{*}{MobileNet-v2\cite{MoV2}}
     & FP & 32/32 & 306.8 & 72.6 \\
     & FP+KD & 32/32 & 306.8 & 73.41\\
    \cline{2-5}
     & DSQ\cite{DSQ} & 4/4 & 15.8 & 64.8 \\
     & LSQ\cite{LSQ} & 4/4 & 15.8 & 70.46\\
     & DJPQ\cite{DJPQ} & MP-BOPs & 7.9 & 69.3\\
     & HAQ\cite{HAQ} & MP-BOPs & 8.3 & 69.5\\
     & DuQ+KD\cite{PROFIT} & 4/4 & 5.3 & 69.86\\
    \cline{2-5}
     & NIPQ-MP-4 & 3.79/4.00 & 8.67  & \textbf{71.52} \\
     & NIPQ-BOPs & 6.19/5.31 & 8.25 & \textbf{72.26}\\
     & NIPQ-BOPs & 5.19/4.24 & 5.37 & \textbf{70.92} \\
     & NIPQ-MP-4+KD & 3.80/4.01 & 8.73  & \textbf{72.16} \\
     & NIPQ-BOPs+KD & 6.21/5.35 & 8.31 & \textbf{72.94}\\
     & NIPQ-BOPs+KD & 5.24/4.24 & 5.34 & \textbf{71.58}\\
    \toprule
    \multirow{10}{*}{MobileNet-v3\cite{Mov3}}
     & FP & 32/32 & 218.7 & 74.52 \\
     & FP+KD & 32/32 & 218.7 & 75.8\\
    \cline{2-5}
     & PACT\cite{PACT} & 4/4 & 3.46 & 67.98 \\
     & PACT+KD\cite{PACT} & 4/4 & 3.46 & 70.16\\
     & DuQ\cite{PROFIT} & 4/4 & 3.46 & 69.50\\
     & DuQ+KD\cite{PROFIT} & 4/4 & 3.46 & 71.01\\
    \cline{2-5}
     & NIPQ-MP-4 & 3.97/3.99 & 7.77  & \textbf{72.41}\\
     & NIPQ-BOPs & 5.02/3.71 & 3.26  & \textbf{70.96} \\
     & NIPQ-MP-4+KD & 4.05/4.00 & 7.96 & \textbf{74.49}\\
     & NIPQ-BOPs+KD & 4.89/3.69 & 3.29 & \textbf{72.41}\\
    \bottomrule
    \end{tabular}
    \end{adjustbox}    
    \caption{Top-1 accuracy (\%) of quantized networks on ImageNet dataset. MP-BOPs represents the mixed-quantization with the bit-operations (BOPs) constraint while MP-N is that with N-bit average bit-width.  '*' denotes the first and last layers remaining 8-bit, and KD denotes the knowledge distillation~\cite{KD}.}
    \label{tab:imagenet_table}
 \end{table}
 \begin{table}[t]  

\begin{adjustbox}{width=\columnwidth}
\begin{tabular}{cccccccc}
\hline
\multirow{2}{*}{} & \multicolumn{7}{c}{Target Bit-width (Weight / Activation)} \\ \cline{2-8} 
& \multicolumn{1}{c}{FP/FP} & 5/8 & 4/8  & \multicolumn{1}{c}{3/8}  & 5/5   & 4/4  & 3/3\\ \hline
DoReFa~\cite{DoReFa} & \multicolumn{1}{c}{0.857} &  0.593 & 0.541 & \multicolumn{1}{c}{0.359} & 0.588 & 0.498 & 0.288  \\
PACT~\cite{PACT}              & \multicolumn{1}{c}{0.857} &            0.845          & 0.835          & \multicolumn{1}{c}{0.806} & 0.838          & 0.811          & 0.708          \\
LSQ~\cite{LSQ}               & \multicolumn{1}{c}{0.857} &           0.85          & 0.843          & \multicolumn{1}{c}{0.815} & 0.843          & 0.823          & 0.761          \\
NIPQ-MP              & \multicolumn{1}{c}{0.857} & \textbf{0.851}  & \textbf{0.848} & \textbf{0.833} & \multicolumn{1}{c}{\textbf{0.848}} & \textbf{0.836} & \textbf{0.801} \\
\hline
\end{tabular}
\end{adjustbox}
\caption{mAP comparison of Yolov5-S~\cite{Yolov5} on the PASCAL VOC~\cite{VOC}.}
\label{tab:yolov5}
\end{table}






Table \ref{tab:imagenet_table} shows top-1 accuracy of the quantized networks. NIPQ shows outstanding results for the optimized but hard to quantize networks such as MobileNet-v2/v3, as well as redundant networks such as ResNet-18. As shown in the table, existing methods are inferior to the proposed method, which achives high accuracy in the same bit-width or bit-operations. These outstanding results come from two factors: first, NIPQ allows the network to converge to a more robust space without STE-induced instability, and second, within the resource budget, the quantization hyper-parameters could be automatically tuned without the intervention of any hand-crafted manipulations. The benefit of these properties is maximized in optimized networks such as MobileNet-v2/v3. Note that when the average precision is constrained, NIPQ tries to increase accuracy with additional operations and vice versa. The automated tuning allows us to quantize the network considering our target goal. 

In addition, we conduct an experiment to quantize the object detection task, which is known to be difficult to quantize. The difficulty is rapidly increased because we apply quantization to the advanced optimized network, Yolov5-S~\cite{Yolov5}. Table \ref{tab:yolov5} shows the comparison of existing quantization studies in the same average bit-width. 
Due to large accuracy loss, existing 4-bit solutions are difficult to use in reality, but NIPQ shows practical, reliable quality in 4-bit precision. The results validate the stability of the NIPQ regardless of the difficulty of the target task. 

Finally, to validate the superiority of NIPQ on the regression application, we apply quantization to the super-resolution task. In the case of super-resolution task, that conducts image restoration, the advantage of dynamic quantization, whose quantization parameters are updated regarding input data, is well demonstrated. Even in this case, NIPQ outmatches the best dynamic method, DDTB~\cite{zhong2022dynamic}. More results are available in the Supplementary Materials.

\section{Conclusion}
To achieve reliable output in low-precision, we propose NIPQ, which enhances the benefit of PQT pipeline by integrating the idea of truncation and providing theoretical support for it. NIPQ automates the layer-wise low-precision optimization for an arbitrary network by updating the network parameters and quantization parameters jointly via gradient descent toward minimizing the target loss. Our extensive results show that the proposed scheme outperforms all of the existing studies by a large margin for various vision and language applications. 

\subsubsection*{Acknowledgements.} This work was supported by IITP grant funded by the Korea government (MSIT, No.2019-0-01906, No.2021-0-00105, and No.2021-0-00310). We appreciate valuable comments from Myeonghwan Ahn at SNU. 

\pagebreak
\begin{center}
\textbf{\large Supplementary Materials}
\end{center}
\setcounter{equation}{0}
\setcounter{figure}{0}
\setcounter{table}{0}
\setcounter{page}{1}
\setcounter{section}{0}

\makeatletter
\renewcommand{\theequation}{S\arabic{equation}}
\renewcommand{\thefigure}{S\arabic{figure}}
\renewcommand{\thetable}{S\arabic{table}}
\renewcommand{\thesection}{S-\Roman{section}}

\section{Overview}
In this supplementary material, we present the details of our implementation and additional experimental results for various tasks and different datasets. We provide the following items:
\begin{itemize}
    \item The detailed implementation of the cost loss function in Section \ref{sec:costloss}.
    \item Detailed Configurations of our experiments in Section \ref{sec:config}
    \item The results of quantization for super resolution task in Section \ref{sec:super-resolution}.
    \item Additional experimental results of object detection on MS-COCO dataset in Section \ref{sec:od-coco}.  
    \item An ablation study on the effect of stochastic rounding in Section \ref{sec:stochastic}.
    \item An ablation study on the effect of late training stage in Section \ref{sec:late-training}.
    \item Experimental results on quantization parameter robustness in Section \ref{sec:qunat-rob}.
    \item The visualization of the quantization noise distribution in Section \ref{sec:quant-noise}
\end{itemize}

\section{Cost Loss Function} \label{sec:costloss}
In order to restrict the utilization of memory and computation resources, we introduce an additional cost loss function in addition to the target loss, as explained in Equation (4) of the main paper. The cost functions for the memory consumption $\mathcal{L}_{cost-MP}$ and the computation cost $\mathcal{L}_{cost-BOP}$ are defined as follows:
\begin{equation}
    \mathcal{L}_{cost-MP} = \lambda_{w}h(\frac{\Sigma_i \lfloor b^w_i \rceil \cdot e^w_i}{\Sigma_i e^w_i} - b_t ) + \lambda_{a}h(\frac{\Sigma_i \lfloor b^a_i \rceil \cdot e^a_i}{\Sigma_i e^a_i} - b_t ),
\end{equation}
\begin{equation}
    \mathcal{L}_{cost-BOP} = \lambda_{b}h(\Sigma_i 
    \lfloor b^w_i \rceil \cdot \lfloor b^a_i \rceil \cdot {OPS}_i - b_t ),
\end{equation}
where $h( \cdot )$ denotes Huber loss, $b^w_i$/$b^a_i$ denote the bit-width of i-th layer's weight/activation, $e^w_i$/$e^a_i$ are the number of elements in the i-th layer's weight/activation, $OPS_{i}$ is FLOPS of the i-th layer and $b_t$ denotes the target bit-width. $\mathcal{L}_{cost-MP}$ regularizes the average bit-width of activation/weight to the target bit, and $\mathcal{L}_{cost-BOP}$ regularizes the sum of overall bit-operation (BOPs) to the target BOPs. Note that we utilize the bit-operations (BOPs) as a representative metric to measure the computation cost of a neural network, which is commonly used in many previous studies~\cite{HAWQ,HAWQv2,HAWQv3}. However, any arbitrary differentiable function can be used as a drop-in replacement for the cost function, and NIPQ automatically optimizes the layer-wise bit-width to the sweet spot. 

On the other hand, while the per-layer (or per-tensor) bit-width also requires rounding operation during forward operation, NIPQ is not applicable for the bit-width because it relies on the statistics of quantization error, but it is improvable to achieve the statistics for the scalar value. To overcome this limitation, we propose to update the bit-width via stochastic rounding with STE (Section \ref{sec:stochastic}).

\begin{table*}[!t!]
\centering
\caption{Fine-tuning configurations of ImageNet classification task. }
\resizebox{\hsize}{!}{
\begin{tabular}{cc|cc|cc|cc|ccc}
\hline
                                                   &          & \multicolumn{2}{c|}{Epoch} & \multicolumn{2}{c|}{SGD} & \multicolumn{2}{c|}{Cosine annealing with warmup} & \multicolumn{3}{c}{$\lambda$} \\ \hline
\multicolumn{2}{c|}{Configuration}                            & Stage-1      & Stage-2     & LR    & Weight decay     & Warmup len           & $\eta_{min}$               & $\lambda_w$  & $\lambda_a$  & $\lambda_b$  \\ \hline
\multicolumn{1}{c|}{ResNet-18}                     & ImageNet & 40           & 3           & 0.04  & $1\times10^{-5}$ & 3                    & $1\times10^{-3}$           & 1            & 1            & 1            \\ \hline
\multicolumn{1}{c|}{\multirow{2}{*}{MobileNet-v2}} & Cifar100 & 30           & 3           & 0.04  & $5\times10^{-5}$ & 5                    & $1\times10^{-3}$           & 1            & 1            & 1            \\
\multicolumn{1}{c|}{}                              & ImageNet & 40           & 3           & 0.04  & $1\times10^{-5}$ & 3                    & $1\times10^{-3}$           & 1            & 1            & 3            \\ \hline
\multicolumn{1}{c|}{MobileNet-v3}                  & ImageNet & 40           & 3           & 0.04  & $1\times10^{-5}$ & 3                    & $1\times10^{-3}$           & 1            & 1            & 3            \\ \hline
\end{tabular}
}
\label{tab:classification_config}

\centering
\caption{Fine-tuning configurations of super-resolution task with EDSR. }
\resizebox{\hsize}{!}{
\begin{tabular}{cc|cc|cc|c|ccc}
\hline
\multicolumn{1}{l}{}           &       & \multicolumn{2}{c|}{Epoch} & \multicolumn{2}{c|}{Adam} & Cosine annealing & \multicolumn{3}{c}{$\lambda$}            \\ \hline
\multicolumn{2}{c|}{Configuration}     & Stage-1      & Stage-2     & LR       & Weight decay   & $\eta_{min}$     & $\lambda_w$ & $\lambda_a$ & $\lambda_b$ \\ \hline
\multicolumn{1}{c|}{EDSR 4bit} & DIV2K & 30           & 10          & $1\times10^{-4}$   & 0            & $1\times10^{-3}$ & 15    & 15        & -    \\ \hline
\multicolumn{1}{c|}{EDSR 3bit} & DIV2K & 40           & 10          & $1\times10^{-4}$   & 0            & $1\times10^{-3}$ & 15    & 15        & -      \\ \hline
\end{tabular}
}
\label{tab:sr_config}

\centering
\caption{Fine-tuning configurations of object detection task with YoloV5-S.}
\resizebox{\hsize}{!}{
\begin{tabular}{c|c|cc|cc|cc|ccc}
\hline
\multicolumn{1}{l}{}          &            & \multicolumn{2}{c|}{Epoch} & \multicolumn{2}{c|}{SGD}    & \multicolumn{2}{c|}{Cosine annealing with warmup} & \multicolumn{3}{c}{$\lambda$} \\ \hline
\multicolumn{2}{c|}{Configuration}         & Stage-1      & Stage-2     & LR     & Weight decay       & Warmup len           & $\eta_{min}$               & $\lambda_w$ & $\lambda_a$ & $\lambda_b$ \\ \hline
\multicolumn{1}{c|}{\multirow{2}{*}{YoloV5-S}} & Pascal VOC & 30           & 5           & 0.0032 & $3.6\times10^{-4}$ & 5                    & $1\times10^{-1}$           & 1           & 1           & -      \\ \cline{2-11}
  & COCO & 35           & 5           & 0.0032 & $3.6\times10^{-4}$ & 5                    & $1\times10^{-1}$           & 1           & 1           & -      \\ \hline
\end{tabular}
}
\label{tab:voc_config}

\centering
\caption{Fine-tuning configurations of GLUE Dataset with BERT-base.}
\resizebox{\hsize}{!}{
\begin{tabular}{cc|cc|cc|cc|ccc}
\hline
\multicolumn{1}{l}{}          &            & \multicolumn{2}{c|}{Epoch} & \multicolumn{2}{c|}{AdamW}    & \multicolumn{2}{c|}{Cosine annealing with warmup} & \multicolumn{3}{c}{$\lambda$} \\ \hline
\multicolumn{2}{c|}{Configuration}         & Stage-1      & Stage-2     & LR     & Weight decay       & Warmup len           & $\eta_{min}$               & $\lambda_w$ & $\lambda_a$ & $\lambda_b$ \\ \hline
\multicolumn{1}{c|}{BERT-base} & GLUE &  25           & 5           & 1e-5 & $1\times10^{-1}$ &5                   & ?       & 1           & 1           & 1     \\ \hline
\end{tabular}
}
\label{tab:bert_config}
\end{table*}

\section{Experimental Configuration} \label{sec:config}
In this paper, all experiments are conducted using GPU servers having 8 x NVIDIA GTX3090 with 24 GB VRAM with 2 x AMD 7313 (16 Core 32 T). The number of GPUs is selected to satisfy the minimum requirement of GPU memory for the target task. All of the experiments are implemented based on the PyTorch \cite{PyTorch} framework (v1.12.1)~\cite{PyTorch}. Our source code is also provided. The additional details of training configuration, e.g., optimizer type, initial learning rate, decay policy, etc., are determined depending on the characteristics of applications and provided in the following paragraphs.

Table \ref{tab:classification_config} shows the configurations of ImageNet training for NIPQ results. In this experiment, we apply quantization to every convolution and linear layer, including the first and last layers. One exception is that the input of the first convolution layer is fixed as 8-bit. We use SGD with momentum optimizer and cosine annealing with warmup scheduling for learning rate adjustment~\cite{SGDR}. $\eta_{min}$ is the final LR multiplier of cosine annealing, and $\lambda_w$, $\lambda_a$, and $\lambda_b$ are the hyper-parameter of resource constraints for the bit-width of weight, bit-width of activation, and BOPs, respectively. When knowledge distillation is triggered, we use EfficientNet-B0~\cite{Eff} as a teacher network. We use the conventional dark-knowledge-based distillation~\cite{KD}. 

Tables \ref{tab:sr_config} and \ref{tab:voc_config} show the detailed configurations of super-resolution task and object detection task, respectively. In both experiments, we keep the precision of the first and last layers as full-precision and apply low-precision quantization to the rest of the layers. In the super-resolution task, we use Adam optimizer~\cite{ADAM} and cosine annealing scheduling for learning rate adjustment. In the object detection task, we use SGD with momentum optimizer and cosine annealing with warmup scheduling for learning rate adjustment. Like the image classification task, $\eta_{min}$ is the final LR multiplier of cosine annealing, and $\lambda_w$, $\lambda_a$, and $\lambda_b$ represent the hyper-parameters of resource constraints for the bit-width of weight, bit-width of activation, and BOPs, respectively. 

Table \ref{tab:bert_config} shows the detailed configurations of BERT-base \cite{devlin2018bert} on the GLUE Task dataset. In this experiment, we modified the code from the huggingface-transformer \cite{wolf-etal-2020-transformers} library. We apply weight quantization to every linear layer except the last classification head. Note that we do not quantize activation or word embedding.
We use the AdamW optimizer and CosineLR scheduler for fine-tuning BERT except for the bit parameters because we find that AdamW can induce instability during training when the magnitude of the cost loss is too large. We use SGD with momentum optimizer for the bit parameters.
Besides, we also find that $\alpha$ and $b$ parameters are not well trained when a single global learning rate is utilized ($1e-5$). For fast and reliable convergence, we use the learning rate of $1e-2$ for bit parameters. In addition, the gradient of $\alpha$ is multiplied $2^b-1$ times over the global learning rate. 


\section{Super Resolution Experiments} \label{sec:super-resolution}


\begin{table}[h]
    \centering
    \begin{adjustbox}{width=0.50\textwidth}
    \begin{tabular}{cccccccccc}
    \toprule
    \multirow{3}{*}{Network} & \multirow{3}{*}{Method} & \multicolumn{8}{c}{Dataset} \\
    \cline{3-10}
    
          &  & \multicolumn{2}{c}{Set5} & \multicolumn{2}{c}{Set14} & \multicolumn{2}{c}{BSD100} & \multicolumn{2}{c}{Urban100}  \\
          \cline{3-10}
           &  & 4bit&3bit & 4bit&3bit & 4bit&3bit & 4bit&3bit  \\
    \midrule
    \multirow{6}{*}{EDSRx2}
          & DoReFa \cite{DoReFa} & 37.22 & 37.13 & 32.82 & 32.73 & 31.63 & 31.57 & 30.17 & 30 \\
          & TFLite \cite{IntArithmetic} & 37.64 & 37.33 & 33.24 & 32.98 & 31.94 & 31.76 & 31.11 & 30.48 \\
          & PACT \cite{PACT} & 37.57 & 37.36 & 33.2 & 32.99 & 31.93 & 31.77 & 31.09 & 30.57 \\
          & PAMS \cite{pams} & 37.67 & 36.76 & 33.2 & 32.5 & 31.94 & 31.38 & 31.1 & 29.5 \\
          & DDTB \cite{zhong2022dynamic} & 37.72 & 37.51 & \textbf{33.35} & 33.17 & \textbf{32.01} & 31.89 & \textbf{31.39} & 31.01 \\
          & NIPQ & \textbf{37.74} & \textbf{37.66} & 33.29 & \textbf{33.20} & \textbf{32.01} & \textbf{31.95} & 31.36 & \textbf{31.13} \\
    \midrule
    \multirow{6}{*}{EDSRx4}
          & DoReFa \cite{DoReFa} & 30.91 & 30.76 & 27.78 & 26.66 & 27.04 & 26.97 & 24.73 & 24.59 \\
          & TFLite \cite{IntArithmetic} & 31.54 & 31.05 & 28.2 & 27.92 & 27.31 & 27.12 & 25.28 & 24.85 \\
          & PACT \cite{PACT} & 31.32 & 30.98 & 28.07 & 27.87 & 27.21 & 27.09 & 25.05 & 24.82 \\
          & PAMS \cite{pams} & 31.59 & 27.25 & 28.2 & 25.24 & 27.32 & 25.38 & 25.32 & 22.76 \\
          & DDTB \cite{zhong2022dynamic} & \textbf{31.85} & 31.52 & \textbf{28.39} & 28.18 & \textbf{27.44} & 27.3 & \textbf{25.69} & 25.33 \\
          & NIPQ & 31.73 & \textbf{31.62} & 28.34 & \textbf{28.25} & 27.41 & \textbf{27.36} & 25.56 & \textbf{25.39} \\
    \bottomrule
    \end{tabular}
    \end{adjustbox}
    \caption{PSNR comparison of quantized EDSR~\cite{EDSR} of scale 4 and scale 2}
    \label{tab:EDSR}
\end{table}

\begin{figure*}[t]
    \small
     \begin{subfigure}{0.19\textwidth}
         \includegraphics[width=\textwidth]{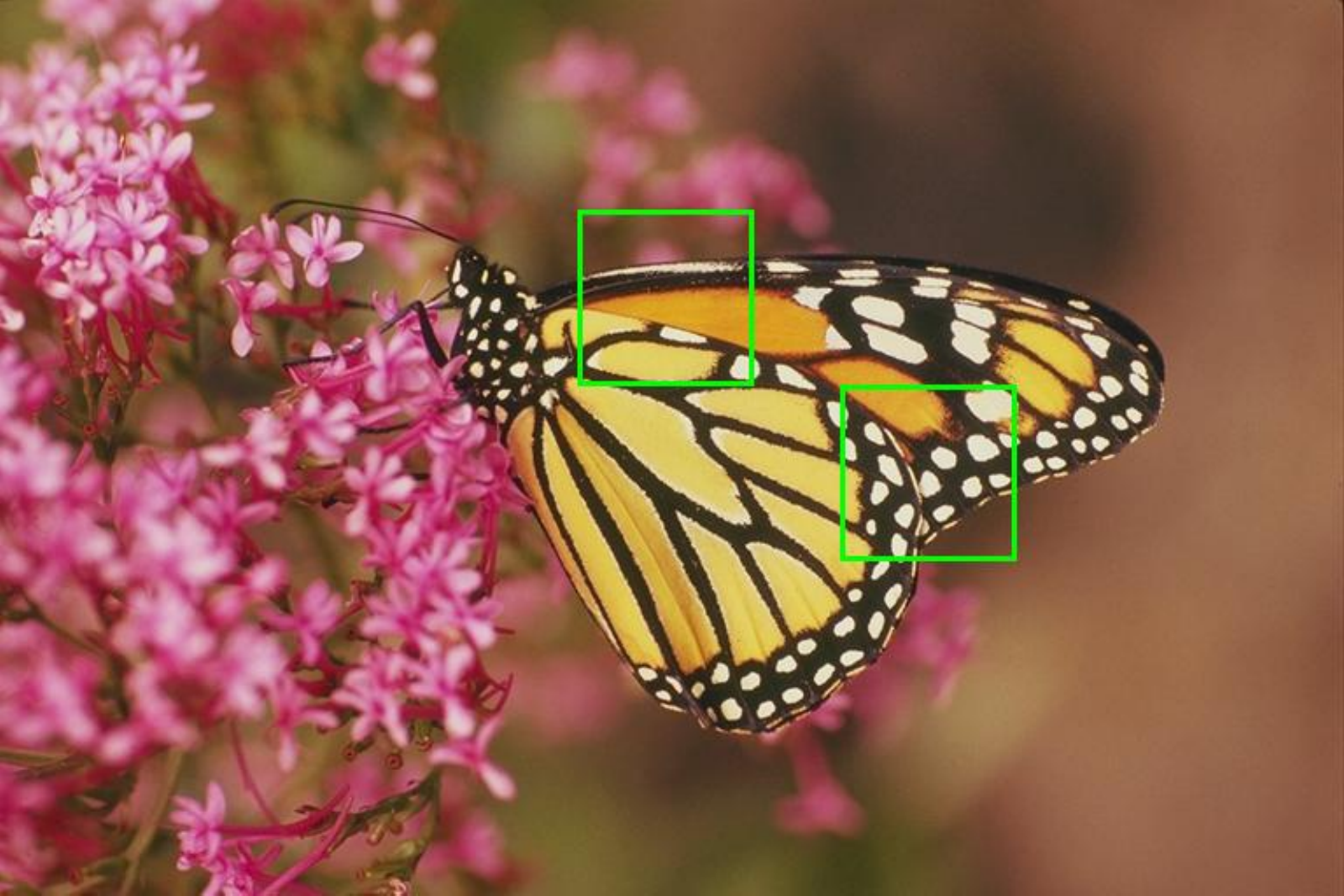}
     \end{subfigure}
     \begin{subfigure}{0.19\textwidth}
         \includegraphics[width=\textwidth]{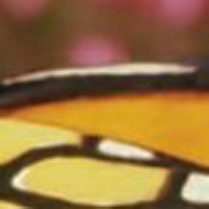}
     \end{subfigure}
     \begin{subfigure}{0.19\textwidth}
         \includegraphics[width=\textwidth]{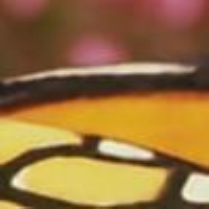}
     \end{subfigure}
     \begin{subfigure}{0.19\textwidth}
         \includegraphics[width=\textwidth]{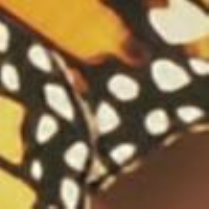}
     \end{subfigure}
     \begin{subfigure}{0.19\textwidth}
         \includegraphics[width=\textwidth]{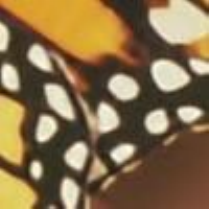}
     \end{subfigure}\\
     \begin{subfigure}{0.19\textwidth}\
         \includegraphics[width=\textwidth]{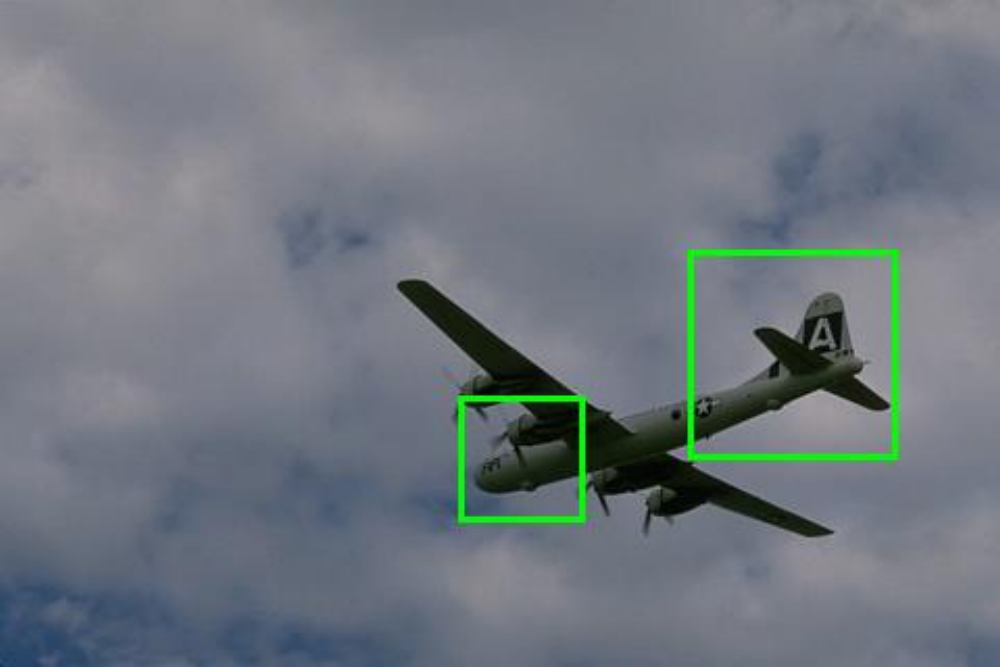}
         \caption{Ground truth}
     \end{subfigure}
     \begin{subfigure}{0.19\textwidth}
         \includegraphics[width=\textwidth]{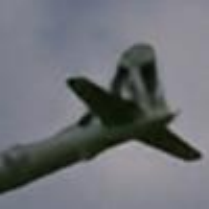}
         \caption{DDTB \cite{zhong2022dynamic}}
     \end{subfigure}
     \begin{subfigure}{0.19\textwidth}
         \includegraphics[width=\textwidth]{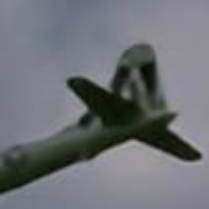}
         \caption{NIPQ}
     \end{subfigure}
     \begin{subfigure}{0.19\textwidth}
         \includegraphics[width=\textwidth]{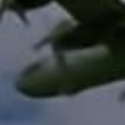}
         \caption{DDTB \cite{zhong2022dynamic}}
     \end{subfigure}
     \begin{subfigure}{0.19\textwidth}
         \includegraphics[width=\textwidth]{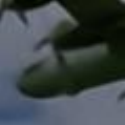}
         \caption{NIPQ}
     \end{subfigure}
     
     \caption{Qualitative results of super resolution on DIV2K dataset. EDSRx4 is quantized into 3-bit for both weights and activations. }
     \label{fig:sr_qual}
\end{figure*}

Table \ref{tab:EDSR} shows the quantitative analysis of NIPQ on super resolution task, and Figure \ref{fig:sr_qual} visualizes the quality of the generated figures. We report PSNR as a quantitative measure, one of the well-known metrics in the area of super resolution. NIPQ outmatches the specialized quantization algorithm for super resolution, DDTB\cite{zhong2022dynamic}, which applies dynamic quantization that adjusts the quantization step size depending on the input data. These experimental results indicate that NIPQ works well in the regression task as well. 

\section{Additional Experiments on Object Detection} \label{sec:od-coco}

\begin{table}[h]
\small
\centering
\caption{mAP comparison of Yolov5-S~\cite{Yolov5} on COCO dataset~\cite{cocodataset}}
\label{tab:yolov5}
\begin{tabular}{c|cccccccccc}
\hline
\multirow{2}{*}{} & \multicolumn{4}{c}{Bit-width (Weight / Activation)}                                                                                                                                                                 \\ \cline{2-5} 
                             & \multicolumn{1}{c|}{FP/FP}   & 5/5            & 4/4            & 3/3            \\ \hline
DoReFa~\cite{DoReFa}         & \multicolumn{1}{c|}{0.354} &          0.266 &          0.24 &           0.191\\
PACT~\cite{PACT}             & \multicolumn{1}{c|}{0.354} &           0.313&           0.294&  0.246        \\
LSQ~\cite{LSQ}               & \multicolumn{1}{c|}{0.354} &           0.32&           0.291&   0.235        \\
NIPQ                         & \multicolumn{1}{c|}{\textbf{0.354}} & \textbf{0.33} & \textbf{0.317} & \textbf{0.284} \\ \hline
\end{tabular}
\end{table}

We conduct an additional experiment on object detection task with the COCO dataset and report mAP on Table \ref{tab:yolov5}. NIPQ obtains the best results compared to existing quantization studies in the same average bit-width.

\begin{figure*}[t]
     \centering
     \begin{subfigure}{0.24\textwidth}
         \centering
         \includegraphics[width=\textwidth]{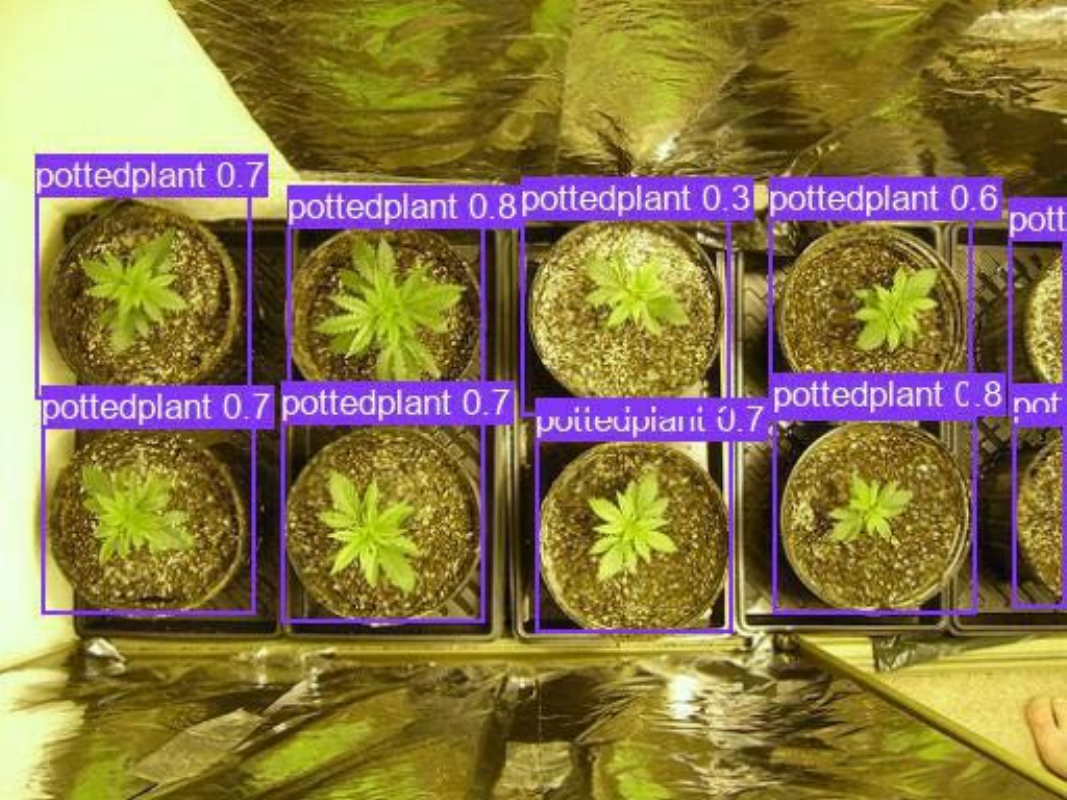}
     \end{subfigure}
     \begin{subfigure}{0.24\textwidth}
         \centering
         \includegraphics[width=\textwidth]{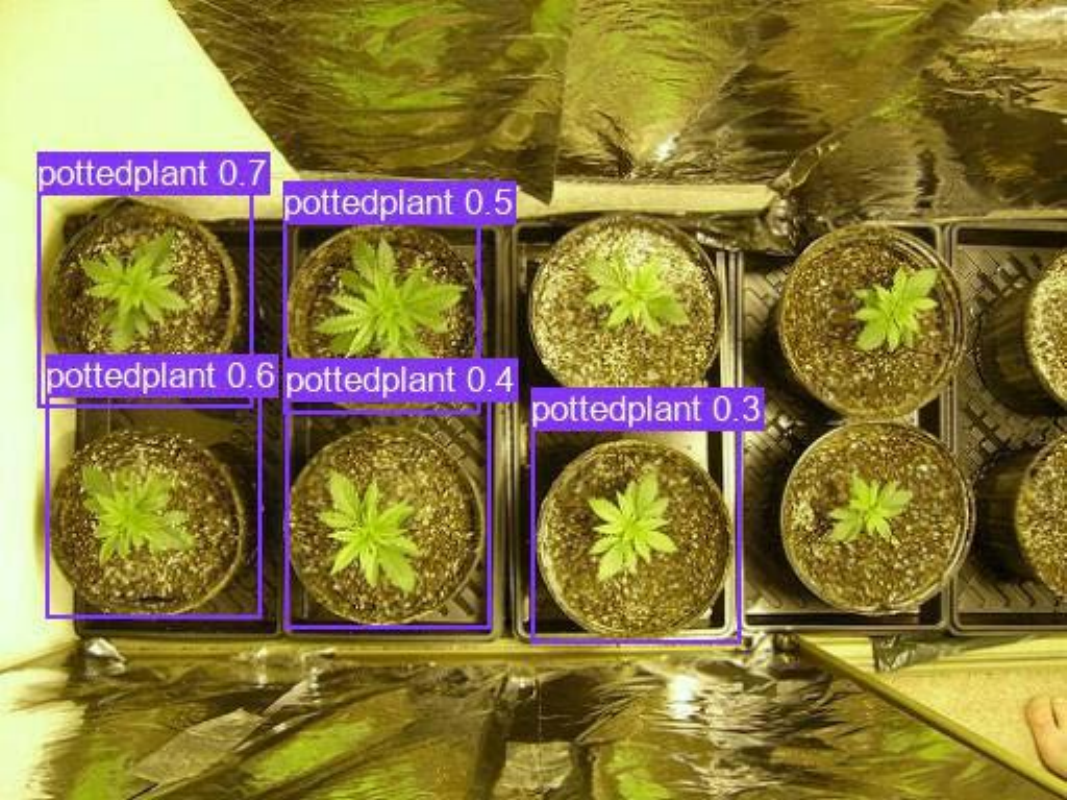}
     \end{subfigure}
     \begin{subfigure}{0.24\textwidth}
         \centering
         \includegraphics[width=\textwidth]{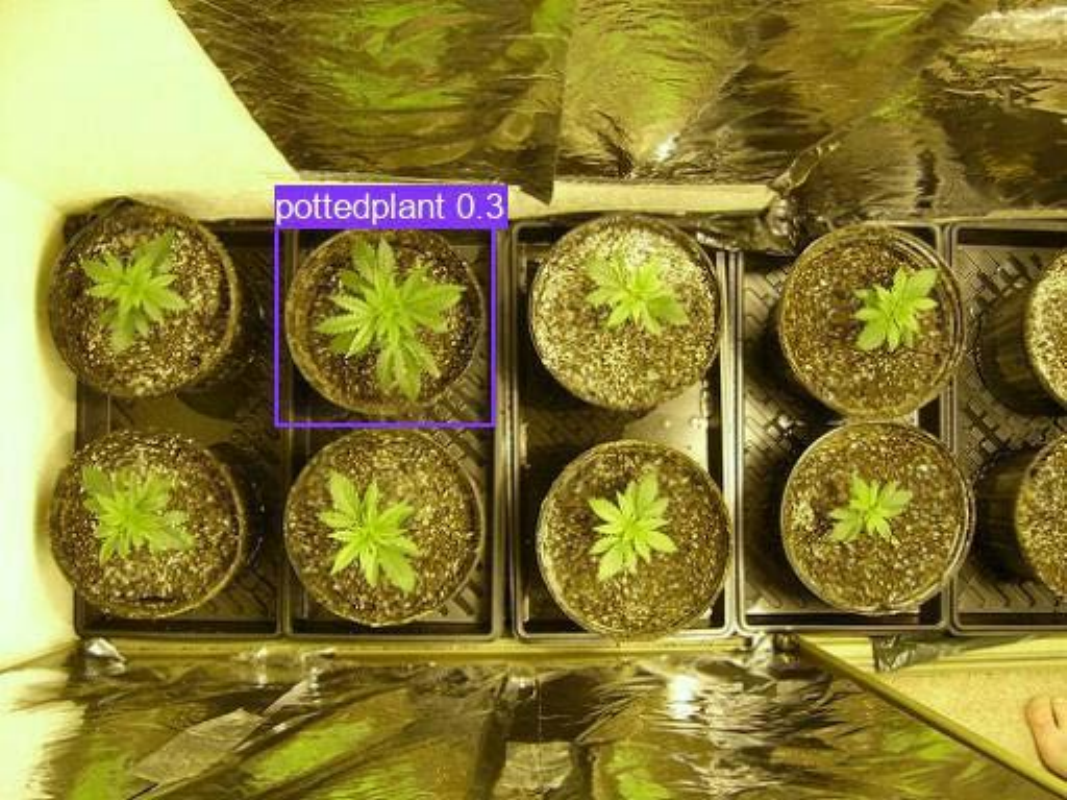}
     \end{subfigure}
     \begin{subfigure}{0.24\textwidth}
         \centering
         \includegraphics[width=\textwidth]{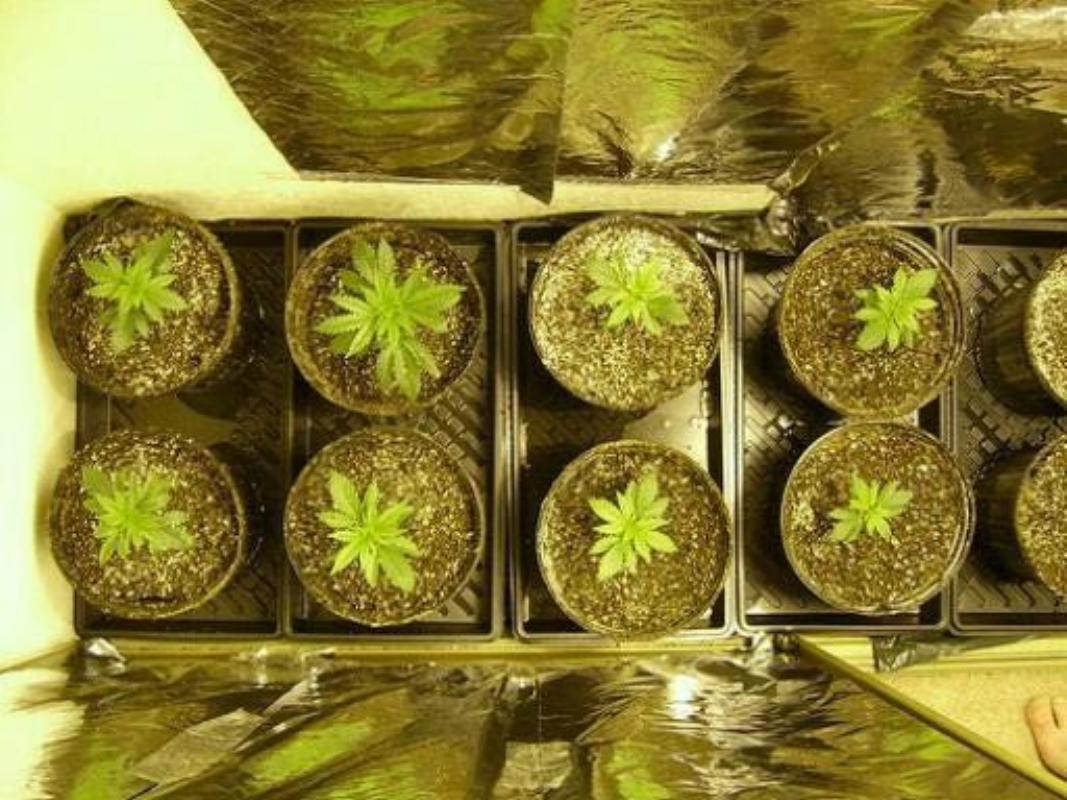}
     \end{subfigure}\\
     \begin{subfigure}{0.24\textwidth}
         \centering
         \includegraphics[width=\textwidth]{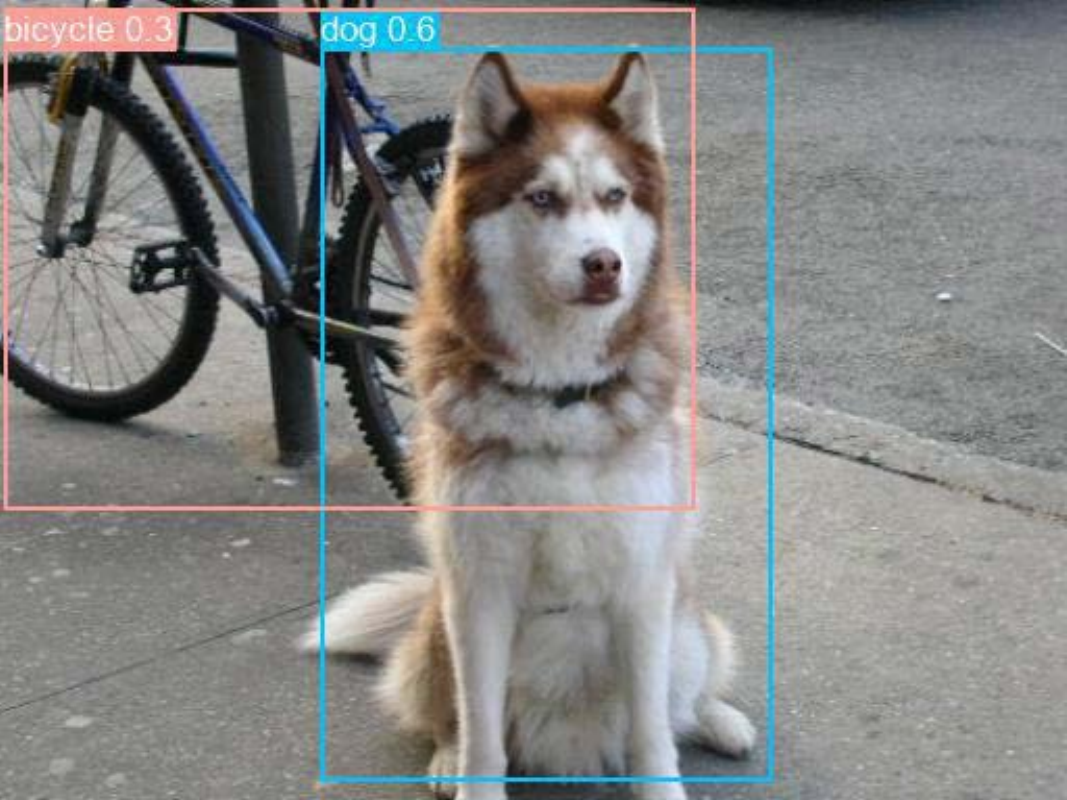}
         \caption{NIPQ}
     \end{subfigure}
     \begin{subfigure}{0.24\textwidth}
         \centering
         \includegraphics[width=\textwidth]{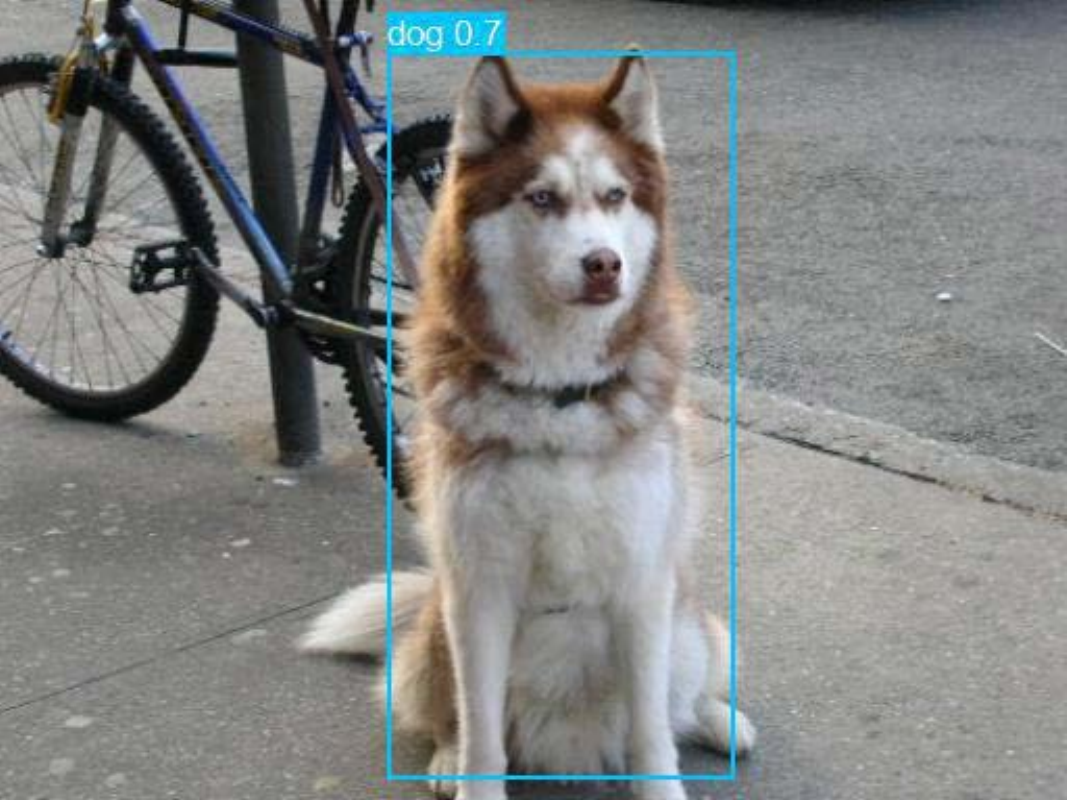}
         \caption{LSQ \cite{LSQ}}
     \end{subfigure}
     \begin{subfigure}{0.24\textwidth}
         \centering
         \includegraphics[width=\textwidth]{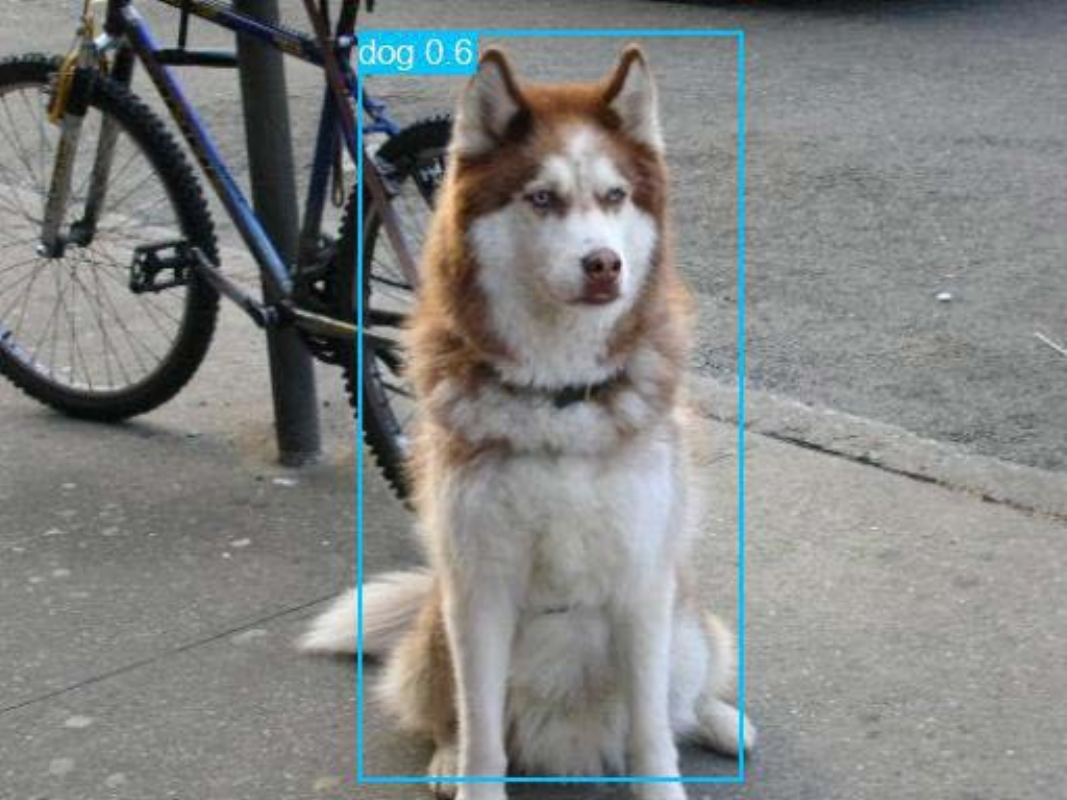}
         \caption{PACT \cite{PACT}}
     \end{subfigure}
     \begin{subfigure}{0.24\textwidth}
         \centering
         \includegraphics[width=\textwidth]{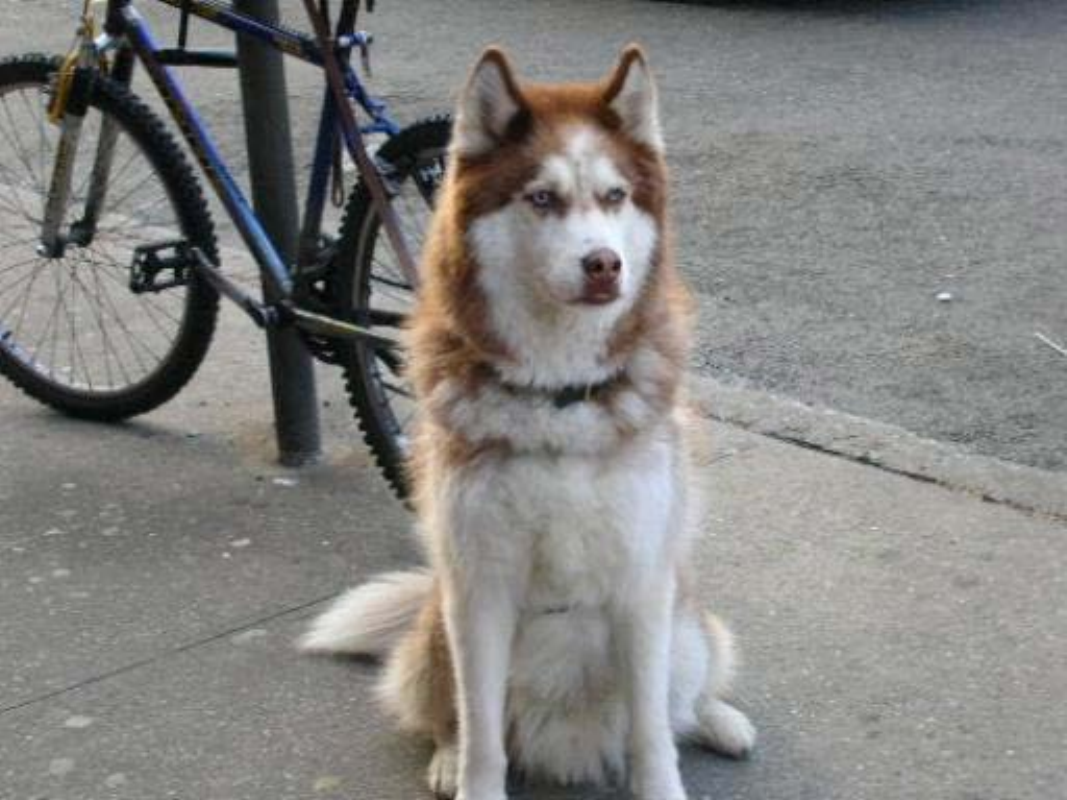}
         \caption{DoReFa \cite{DoReFa}}
     \end{subfigure}
     
     \caption{Qualitative results of object detection on the VOC dataset. Yolov5-S is quantized into 3-bit weights and activations according to each quantization method.}
     \label{fig:pascal_od}
\end{figure*}

In addition, in Figure \ref{fig:pascal_od}, we visualize the qualitative results of NIPQ on the VOC dataset. Bounding box regression and classification results of the quantized network are presented. As shown in the figure, NIPQ works surprisingly well in the 3-bit domain on the challenging object detection problem. YoloV5-S has a complicated structure, and the sensitivity of each layer is highly different. Because NIPQ has the ability to allocate the bit-width aware of the sensitivity automatically and enable stable convergence without STE instability, the quality of the quantized network outperforms all of the previous methods by a large margin.

\section{Stochastic Rounding for Bit-width} \label{sec:stochastic}

\begin{figure}[h]
    \centering
    \includegraphics[width=1.0\columnwidth]{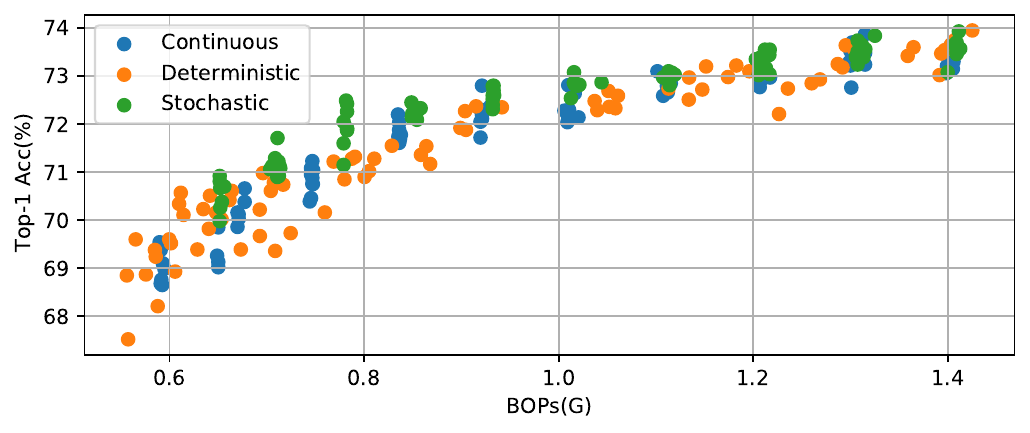}
    \caption{Accuracy comparison of MobileNet-v2 at CIFAR-100 dataset with different bit-width training strategies. We ran the experiments with 10 repetitions, increasing from 0.5 BOPs to 1.4 BOPs by 0.1 BOPs steps.}
    \label{fig:stochastic}
\end{figure}


While we propose an alternative training scheme for quantization instead of using STE, updating the bit-width is a remaining problem that is not addressed in the NIPQ pipeline. The proposed noise proxy is designed to update the learnable parameters by emulating the quantization operator based on PQN. However, the bit-width is assigned as a scalar value per the target tensor, and thereby it is impossible to aggregate the coarse-grained effect of the quantization operator. When we use rounding-based QAT with STE approximation, the bit-width also suffers from the instability of STE, resulting in highly unreliable result, as shown in Figure \ref{fig:stochastic}. Due to this limitation, many previous studies rely on the continuous approximation of bit-width during training~\cite{DJPQ,DiffQ} to avoid the instability problem. However, the representation mismatches to the domain of bit-width, resulting in suboptimal convergence in practice, especially when the target bit-width is in a sub-4-bit domain. In this paper, we propose an alternative idea to utilize the stochastic rounding of bit-width during training. Stochastic rounding is an unbiased estimator, so the learnable bit-width converges to the optimal point as the learning progresses. In addition, the bit-width is evaluated in the discrete domain during training, which mitigates the domain gap between training and inference. As shown in Figure \ref{fig:stochastic}, the stochastic rounding consistently draws the pareto-front line with small variance, which enables us to search for the best quantization configurations within the given resource budget. 

\begin{figure*}[t!]
     \centering
     \begin{subfigure}{0.325\textwidth}
         \centering
         \includegraphics[width=\textwidth]{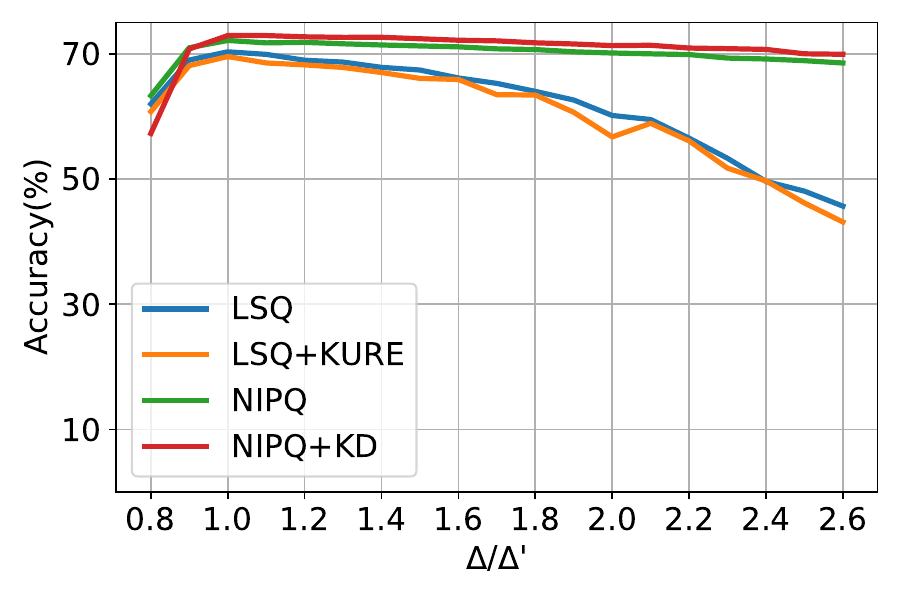}
         \caption{Activation}
     \end{subfigure}
     \begin{subfigure}{0.322\textwidth}
         \centering
         \includegraphics[width=\textwidth]{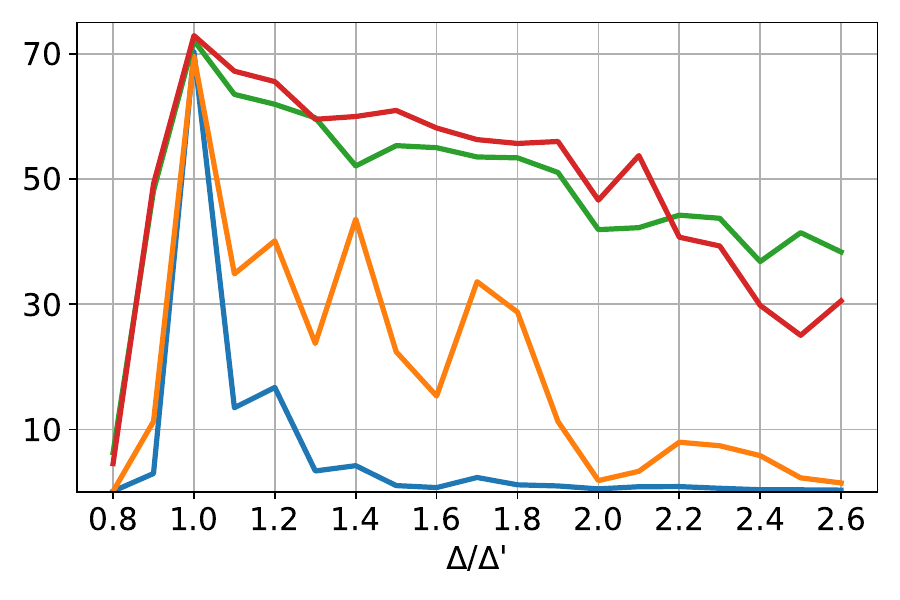}
         \caption{Weight}
     \end{subfigure}
     \begin{subfigure}{0.322\textwidth}
         \centering
         \includegraphics[width=\textwidth]{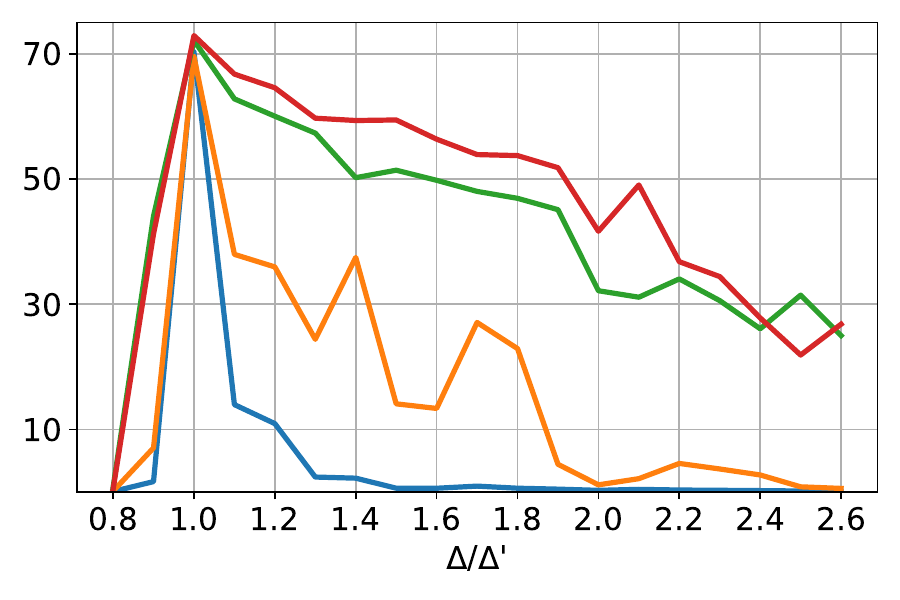}
         \caption{Activation \& Weight}
     \end{subfigure}
     \caption{Robustness of quantized MobileNet-V2 on ImageNet against the change of $\alpha$ of the quantization operator for weight and activation. $\Delta'$ is the trained $\alpha$ and $\Delta$ is the scaled one. NIPQ+KD represents the quantized network with knowledge distillation~\cite{KD2} using EfficientNet-B0~\cite{Eff} as a teacher.}
     \label{fig:robust}
\end{figure*}

\begin{figure*}[t]
     \centering
     \begin{subfigure}{0.5\columnwidth}
         \centering
         \includegraphics[width=\textwidth]{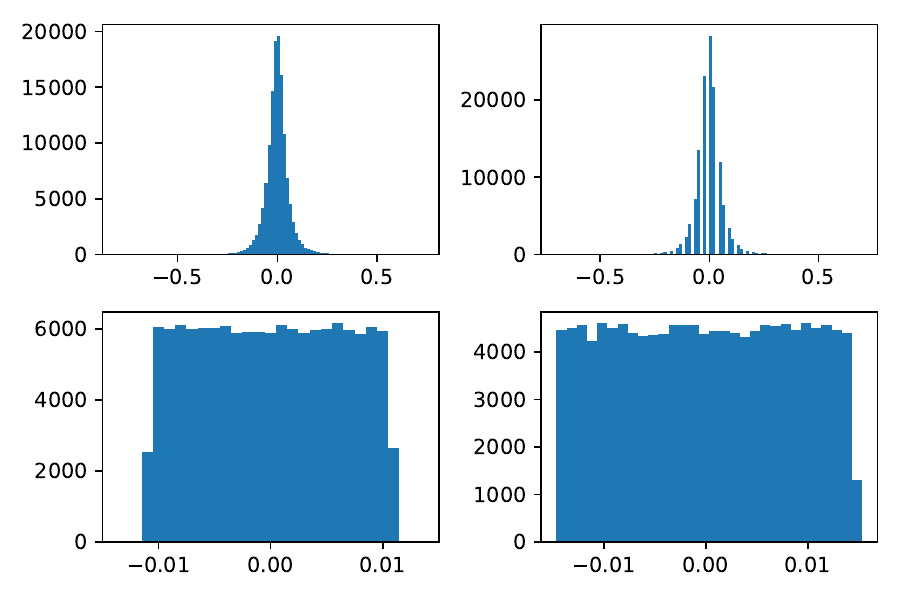}
         \caption{5-bit}
     \end{subfigure}
     \begin{subfigure}{0.5\columnwidth}
         \centering
         \includegraphics[width=\textwidth]{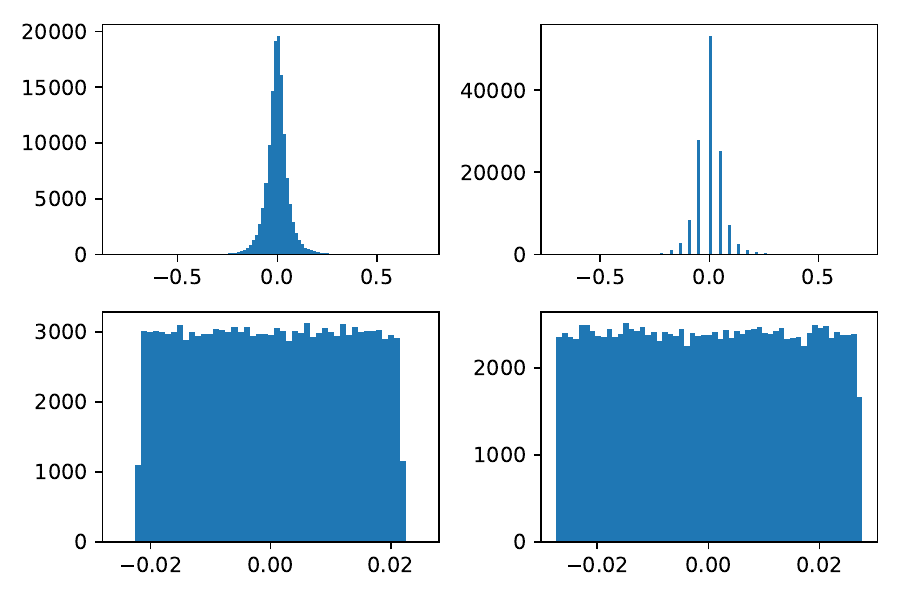}
         \caption{4-bit}
     \end{subfigure}
     \begin{subfigure}{0.5\columnwidth}
         \centering
         \includegraphics[width=\textwidth]{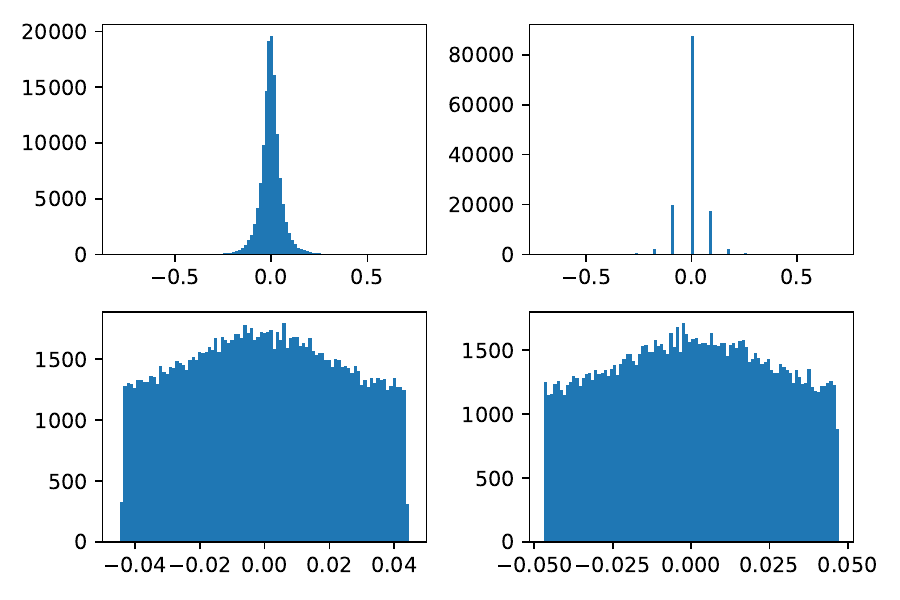}
         \caption{3-bit}
     \end{subfigure}
     \begin{subfigure}{0.5\columnwidth}
         \centering
         \includegraphics[width=\textwidth]{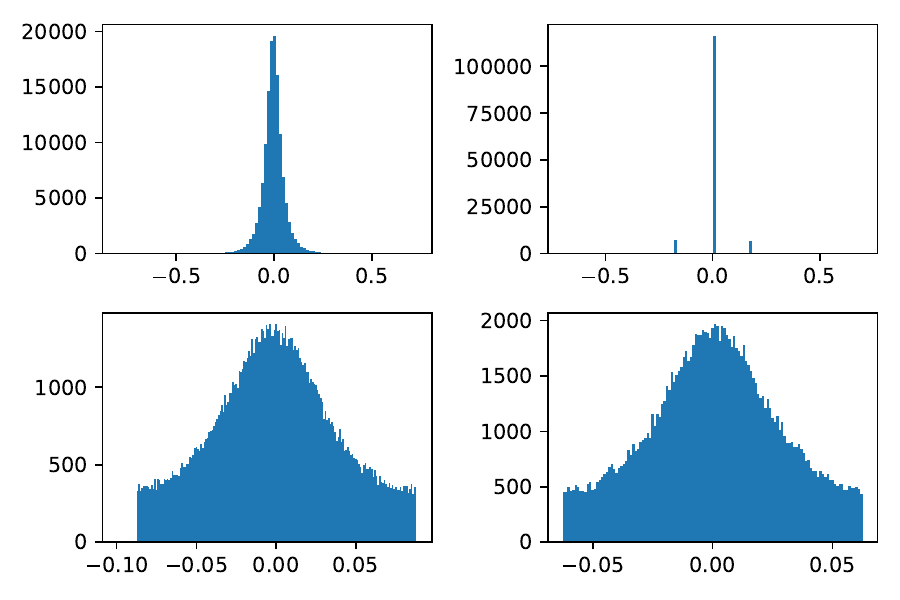}
         \caption{2-bit}
     \end{subfigure}
     \caption{Quantization noise distribution of
     ResNet-18's 12-th convolution weight. (left top) 32 bit (FP) distribution. (right top) N-bit uniform quantized distribution. (left bottom) Real quantization noise distribution. (right bottom) Sampled quantization noise distribution.}
     \label{fig:noise}
\end{figure*}

\section{Comparison for the Late Training Stage} \label{sec:late-training}

\begin{table}[h] \
\centering
\small
\caption{Comparison of accuracy regarding the late training stage. MobileNet-v2 is trained in 30 epochs and finetuned in 3 epochs on the CIFAR-100 dataset. The target computation overhead is 1.0 GBlops  }
\begin{adjustbox}{width=0.48\textwidth}
\begin{tabular}{cc|c|ccc}
\hline
 \multicolumn{2}{c|}{}& FP & Without Tuning & BN update & QAT finetune \\
 \cline{3-6}
 \multicolumn{2}{c|}{Top-1} & 75.04 & 70.45 & 72.99 & 73.29 \\
 \hline
\end{tabular}
\end{adjustbox}
\label{tab:LTtable}
\end{table}

In Table \ref{tab:LTtable}, we show the results of NIPQ with different late training stage policies. As shown in the table, BN update offers a large performance benefit compared to the accuracy of the NIPQ training without the late stage tuning. Because PQN of NIPQ disturbs the statistics of normalization layers, the correction of the statistics is essential to maximize the accuracy in the inference phase. In addition, QAT finetune offers an additional performance improvement by giving an additional chance to adjust the learnable parameters of the entire network without the effect of PQN with a small learning rate, which enables the stabilization of network parameters near the optimal point with the tiniest effect of STE instability.

\section{Robustness of the quantization parameters} \label{sec:qunat-rob}

NIPQ also enhances the robustness of the quantization parameters as well as the network parameter. Figure \ref{fig:robust} visualizes the results of measuring the accuracy while changing the quantization step size or the truncation interval. The more robust the network, the more it can endure the change of the quantization configuration. As shown in the figure, NIPQ shows comparable or superior results to the previous best algorithm for robustness, KURE \cite{KURE}. It is especially worthy that existing studies have focused on improving the robustness of weight only~\cite{Grad,KURE}, but NIPQ also improves the robustness of activation as well by a large margin. To the best of our knowledge, this is for the first time that activation robustness can be improved, which is a crucial benefit of deploying networks in a noisy environment. 

\section{Quantization Noise Distribution} \label{sec:quant-noise}


In Figure \ref{fig:noise}, we visualize the quantization noise distribution of ResNet-18's 12-th convolution weight in different bit-widths. When applying quantization, not only rounding but also truncation is applied. The previous study argues that the quantization noise distribution follows a uniform distribution regardless of input distribution when the number of bits is sufficiently large \cite{PQN}. According to our observation, the statement is held in practice when the bit-width is larger than 4-bit. However, in sub-4-bit precision, the distribution of noise seems to follow a bell-shaped curve instead of a uniform distribution. Due to these characteristics, conventionally uniform or gaussian distributions are often used to approximate PQN~\cite{DJPQ,DiffQ}. However, as presented in this paper, the precise sampling of PQN following the quantization error distribution is essential to guarantee the convergence on the optimal point, while the uniform distribution shows comparable results in practice empirically. In this work, we realize the sampling process of quantization error distribution on GPU with practical performance as follows: first, the probability density function (PDF) of the quantization error distribution is estimated based on the histogram with 256 bins. Then, the distribution is sampled from the estimated PDF of the histogram. As shown in Figure \ref{fig:noise}, the sampled distribution precisely follows the quantization error distribution.

{\small
\bibliographystyle{ieee_fullname}
\bibliography{egbib}
}






\end{document}